\documentclass[10pt,journal,compsoc]{IEEEtran}
\usepackage{xcolor}         

\ifCLASSOPTIONcompsoc
  \usepackage[nocompress]{cite}
\else
  \usepackage{cite}
\fi

\ifCLASSINFOpdf
\else
\fi

\usepackage{url}            
\usepackage{booktabs}       
\usepackage{amsfonts}       
\usepackage{nicefrac}       
\usepackage{microtype}      
\usepackage{times}
\usepackage{epsfig}
\usepackage{graphicx}
\usepackage{amsmath}
\usepackage{amssymb}
\usepackage{subfiles}
\usepackage{multirow}
\usepackage{booktabs}
\usepackage{xcolor}
\usepackage{subcaption}
\usepackage{amssymb}
\usepackage{pifont}
\newcommand{\cmark}{{\color{blue}\ding{51}}}
\newcommand{\xmark}{{\color{red}\ding{55}}}
\usepackage{hyperref}
\hypersetup{
    colorlinks = true
}
\usepackage{bbm}
\newcommand\indicator[1]{\mathbbm{1}[#1]}
\def\CircleArrowright{\ensuremath{%
  \rotatebox[origin=c]{310}{$\circlearrowright$}}}

\begin{document}
%
\title{Deep Learning for Embodied Visual Navigation Research: A Survey}

\author{Fengda Zhu, Yi Zhu, Vincent CS Lee, Xiaodan Liang and Xiaojun Chang
\IEEEcompsocitemizethanks{\IEEEcompsocthanksitem Fengda Zhu and Vincent CS Lee are with Department of Data Science and AI, Faculty of Information Technology, Monash University.
\IEEEcompsocthanksitem Yi Zhu is with University of Chinese Academy of Sciences.
\IEEEcompsocthanksitem Xiaodan Liang is with School of Intelligent Systems Engineering, Sun Yat-sen University. 
\IEEEcompsocthanksitem Xiaojun Chang is with School of Computing Technologies, RMIT University.
}
\thanks{Manuscript received July 8, 2021.}
}


\IEEEtitleabstractindextext{%
\begin{abstract}
``Embodied visual navigation'' problem requires an agent to navigate in a 3D environment mainly rely on its first-person observation. 
This problem has attracted rising attention in recent years due to its wide application in vacuum cleaner, and rescue robot, etc. 
A navigation agent is supposed to have various intelligent skills, such as  visual perceiving, mapping, planning, exploring and reasoning, etc. 
Building such an agent that observes, thinks, and acts is a key to real intelligence. 
The remarkable learning ability of deep learning methods empowered the agents to accomplish embodied visual navigation tasks.
Despite this, embodied visual navigation is still in its infancy since a lot of advanced skills are required, including perceiving partially observed visual input, exploring unseen areas, memorizing and modeling seen scenarios, understanding cross-modal instructions, and adapting to a new environment, etc. 
Recently, embodied visual navigation has attracted rising attention of the community, and numerous works has been proposed to learn these skills. 
This paper attempts to establish an outline of the current works in the field of embodied visual navigation by providing a comprehensive literature survey. We summarize the benchmarks and metrics, review different methods, analysis the challenges, and highlight the state-of-the-art methods. 
Finally, we discuss unresolved challenges in the field of embodied visual navigation and give promising directions in pursuing future research. 
\end{abstract}

}

\maketitle

\IEEEdisplaynontitleabstractindextext

%
\IEEEpeerreviewmaketitle

\IEEEraisesectionheading{\section{Introduction}\label{sec:introduction}}

\begin{figure*}[t]
	\centering
	\includegraphics[width=0.99\linewidth]{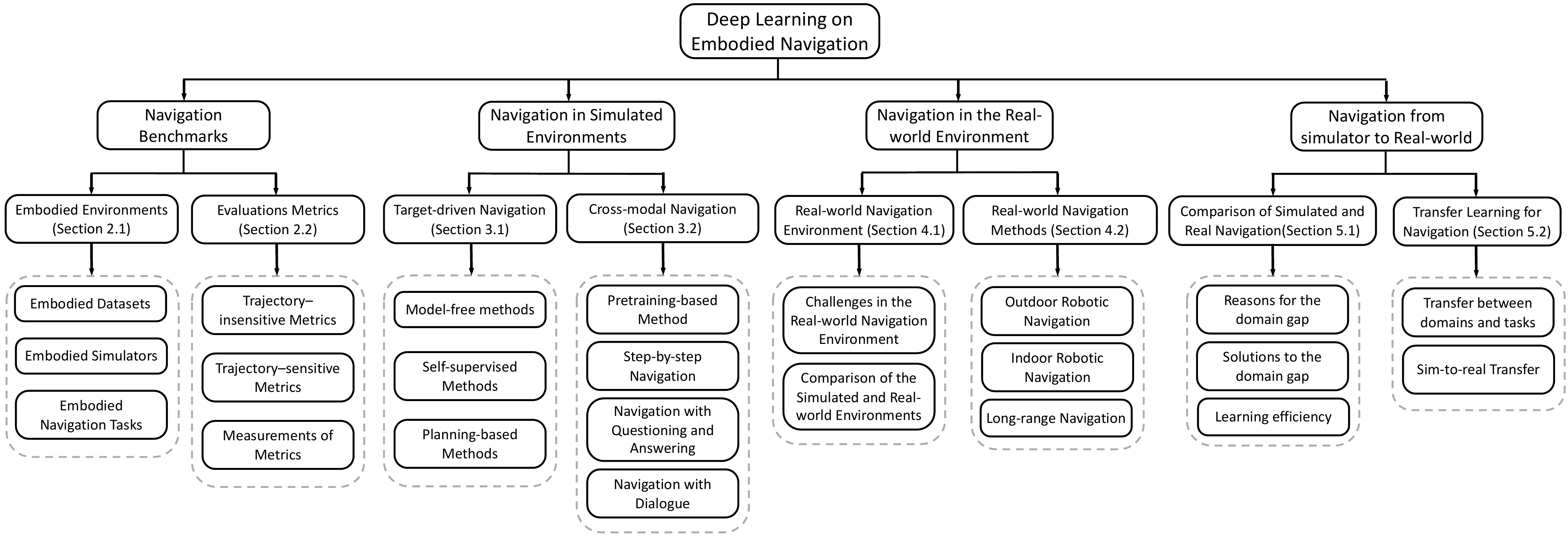}
	\caption{
	A taxonomy of deep learning methods for embodied navigation. }
	\label{fig:outline}
	\vspace{-8pt}
\end{figure*}

\begin{figure}[t]
	\centering
	\includegraphics[width=0.72\linewidth]{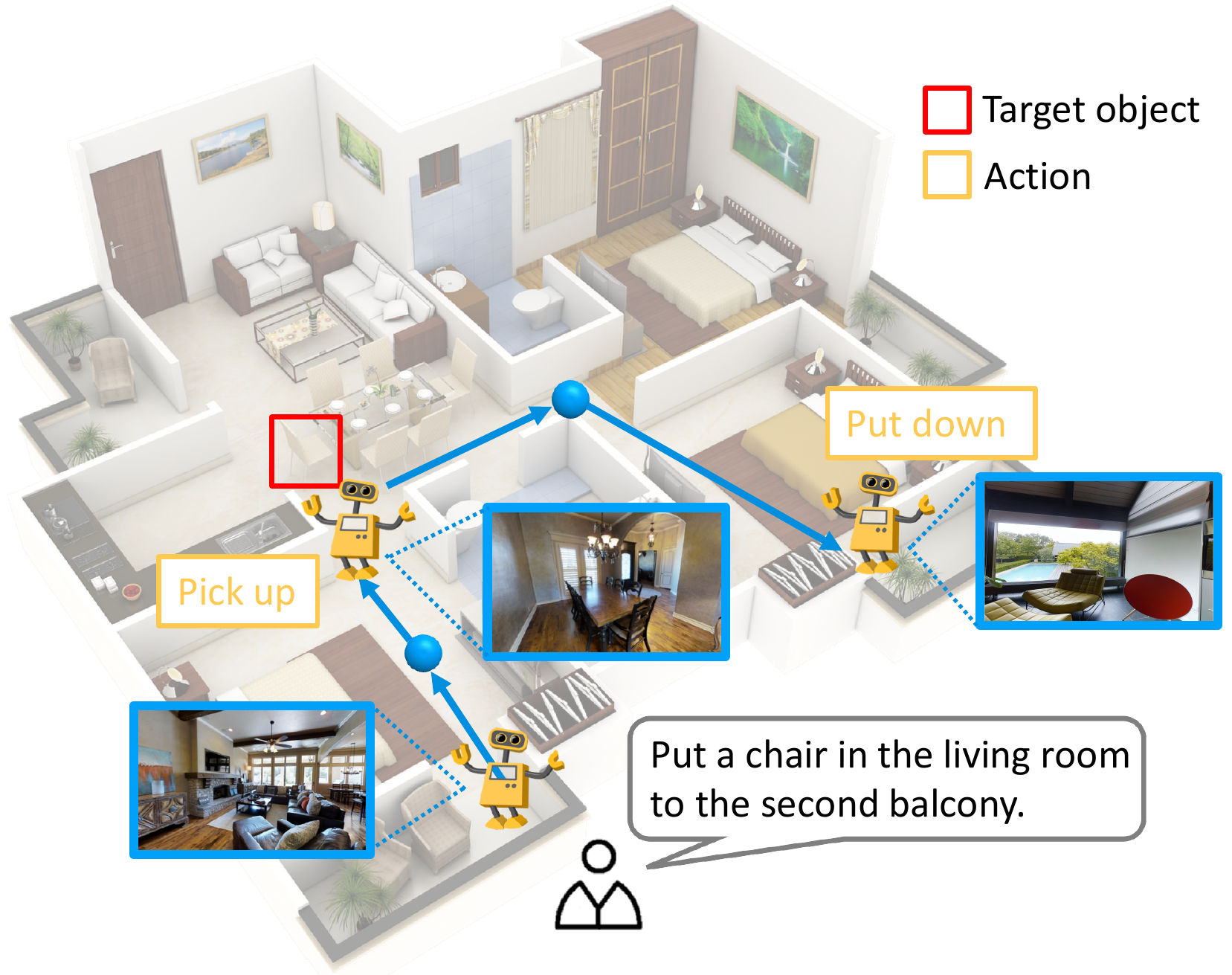}
	\caption{
	A demonstration of a navigation process, in which a robot move to several places to accomplish a task. }
	\label{fig:demo}
	\vspace{-8pt}
\end{figure}

\IEEEPARstart{B}{uilding} a robot to accomplish tasks autonomously in place of humans has been a topic researched for a long time~\cite{krizhevsky2017imagenet, lecun2015deep, thrun2005probabilistic}. 
Some complex applications, such as vacuum cleaning, disabled helping and rescuing, require an agent to navigate to finish several sub-tasks in different places in a 3D embodied environment. 
Therefore, navigation is one of the key capabilities in building the intelligent navigable robots in the real world. 
During the process of navigation, a robot needs to move around to find the target location by perceiving embodied visual inputs, which is named as ``embodied visual navigation''. 
The agent that interacts with the environment through its physical entity within that environment is named ``embodied agent''~\cite{dourish2001where}. 
Fig.~\ref{fig:demo} demonstrates a navigation process. An agent firstly receives an instruction ``\emph{Put the chair in the living room into the second balcony}''. Then it navigates to find the target chair. The agent picks up the chair and navigate to the balcony and put it down. 

Early works on robotics navigation~\cite{kruse2013human, desouza2002vision} mainly rely on hand-crafted features like optical flow
and traditional algorithms like Markov localization~\cite{thrun2000probabilistic}, incremental localization~\cite{kosaka1992fast}, or landmark tracking~\cite{kabuka1987position}. 
These methods involve lots of hyper-parameters and cannot generalize well in unseen environments. 
Recent developments of deep learning reveal its ability to learn a robust model from large-scale data. 
Vision robots trained by end-to-end deep learning methods is more robust, have less hyper-parameters, and have better generalization ability in unseen environments.

However, some challenges are going to be tackled in achieving deep learning for embodied visual navigation:
1) collecting data from the real world is expensive;
2) the model learned from partial observation is unstable;
3) it is difficult to learn the skills for long-term navigation such as exploration and memorization;
4) perceiving natural language instructions is challenging because natural language is diverse with flexible formats;
5) the large domain gap between the simulated environment and the real-world environment impedes the adaptation of navigation policy, etc.  

This paper discusses the related works in robot navigation and give a promising direction in  building real-world navigation robots. 
The structure of this paper is shown in Fig.~\ref{fig:outline}.
Training and testing in the real-world has many disadvantages: 
1) The data sampling efficiency in the real world is very low since a real robot can only sample a trajectory at once while a simulator can efficiently sample trajectories in multi-processing; 
2) the complexity and the dynamic of the real-world environment hinges the reproductivity; 
3) there is large domain bias between different environments, etc. 
The development of 3D simulation technology enables researchers to construct a simulated environment~\cite{wu2018building, savva2017minos, anderson2018vision, savva2019habitat, xia2018gibson} to study in building a robust navigation agent within it. 
Simulators render these 3D assets to generate RGB-D images and provide sensors like physical sensors, GPS sensors to simulate a realistic embodied robotic environment. 
Learning to navigation in a simulated environment is a broad field with lots of challenges to solve. 
In solving target-driven navigation problem, researchers propose 
model-free methods~\cite{zhu2017target, wu2018building, mousavian2019visual}, 
self-supervised methods~\cite{xie2016model, jaderberg2017reinforcement, mirowski2017learning}, 
planning-based methods~\cite{gupta2020cognitive, qi2020learning, chaplot2020learning}. 
Perceiving natural language is a challenging task due to its diversity and complexity. 
It requires the agents not only can following a sentence instruction step-by-step~\cite{anderson2018vision, wang2018look, fried2018speaker}, but also understand dialogues~\cite{gupta2020cognitive, qi2020learning, chaplot2020learning} or navigate to answer questions~\cite{gupta2020cognitive, qi2020learning, chaplot2020learning}. 
In building a real-world navigation robots, some works~\cite{jackel2006darpa, morad2021embodied, thorpe1988vision} proposed to train an agent in real-world environments directly while other works~\cite{traor2019discorl, huang2019transferable, yan2020multimodal} propose to introduce transfer learning to transfer the learned navigation policy from simulated environments to the real-world environment.

Compared with previous surveys of robotic navigation~\cite{desouza2002vision, kruse2013human}, 
our paper focuses on deep learning methods that solve the embodied navigation problems: 
\begin{enumerate}
  \item To the best of our knowledge, our paper is the first comprehensive study on the advances of deep learning methods on embodied navigation tasks. 
  \item This paper summarizes and compares their unique insights of recently proposed embodied navigation datasets, simulators and tasks. 
  \item This paper introduce deep learning methods for embodied visual navigation, including their motivations and contributions. 
  \item This paper classifies the research results in recent years, and gives some promising embodied navigation directions. 
\end{enumerate}

The paper is organized as follows. 
Sec.~\ref{sec:embodied_navigation_environments} discusses the current embodied datasets and embodied simulators. 
Sec.~\ref{sec:embodied_navigation_benchmarks} introduces the embodied navigation benchmarks including the navigation tasks and navigation metrics. 
Sec.~\ref{sec:methods_simulated_env} lists the methods to train an agent navigation in a simulated embodied navigation environment, where Sec.~\ref{sec:target_navigation} lists the methods for target-driven tasks and Sec.~\ref{sec:navigation_with_natural_language} introduces the methods for cross-modal tasks. 
Sec.~\ref{sec:real_world_embodied_navigation} summarizes the works that builds a navigation robot. 
Sec.~\ref{sec:challenge_solution_real_navigation} 
illustrates the domain gap betweeen simulated environments and the real-world environment, and introduce the methods that solves these challenges. 
In Sec.~\ref{sec:future_direction}, we highlight the recent state-of-the-art works, discuss about the limitations of current works, and propose promising directions in building a real-world navigation robot.

\begin{table*}
	\centering
	\resizebox{1.0\textwidth}{!}{
	\setlength{\tabcolsep}{0.9em}
    {\renewcommand{\arraystretch}{1.05}
	\begin{tabular}{|l|c|ccc|ccc|}
		\hline
        Dataset & Year & Scenes & Rooms & Object Catagories & RGB & Depth & 2D Semantics \\ 
		\hline 
		Stanford Scene*~\cite{fisher2012example} & 2012 & 130 & 130 & - & synthetic & \xmark & \xmark \\
		SceneNet*~\cite{handa2016scenenet} & 2016 & 57 & 57 & 218 & synthetic & \xmark & \cmark \\
		2D-3D-S*~\cite{armeni2017joint} & 2017 & 270 & 270 & 13 & synthetic & \cmark & \cmark \\
		SUNCG~\cite{song2017semantic} & 2017 & 45,622 & 775,574 & 84 & synthetic & \cmark & \cmark \\
		CHALET~\cite{yan2018chalet} & 2018 & 10 & 58 & 150 & synthetic & \xmark & \xmark \\
		Matterport3D~\cite{chang2017matterport3d} & 2017 & 90 & 2,056 & 40 & realistic & \cmark & \cmark \\
		Gibson~\cite{xia2018gibson} & 2018 & 572 & 8,854 & 84 & realistic & \cmark & \cmark \\
		Replica~\cite{straub2019the} & 2019 & 18 & 35 & 88 & realistic & \cmark & \cmark \\
		\hline
	\end{tabular} 
	}
	}
	\caption {Comparison of existing embodied datasets (*: the datasets render only a room as scene). }
	\label{table:datasets}
	\vspace{-10pt}
\end{table*}

\begin{figure}[t]
	\centering
	\includegraphics[width=0.99\linewidth]{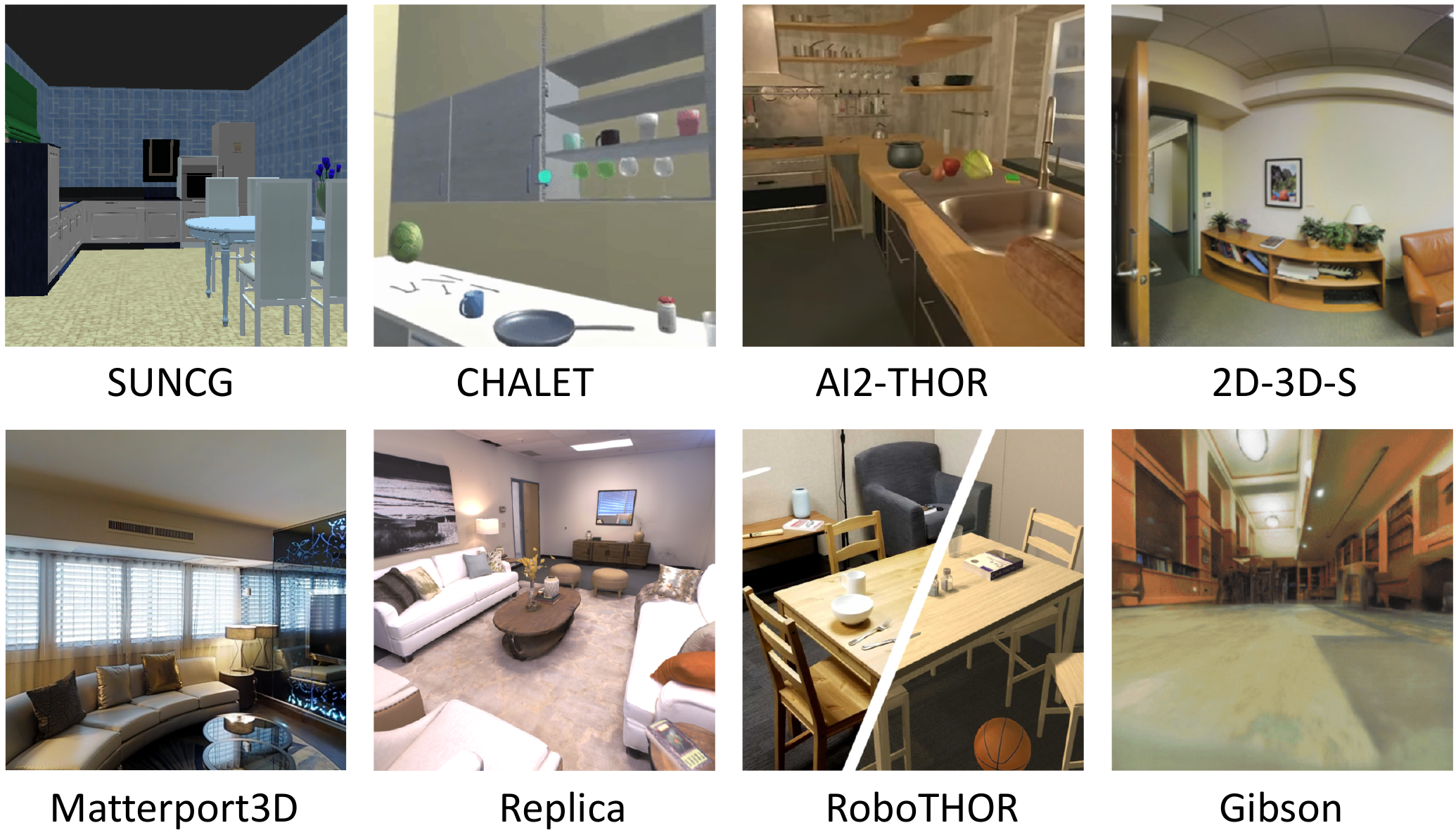}
	\caption{
	The render scenes of each dataset. }
	\label{fig:datasets}
	\vspace{-8pt}
\end{figure}


\section{Embodied Navigation Environments}
\label{sec:embodied_navigation_environments}

Here, we discuss the environments used for embodied navigation. We summarize the dataset that provides 3D assets and the simulators that render assets and provide interactive interfaces for navigation agents. 

\subsection{Embodied Datasets}
\label{sec:embodied_datasets}
An embodied dataset contains 3D assets like textures and meshes for rendering and other configuration data like object location, object category and camera pose for high-level tasks. 
A comparison of the proposed datasets is shown in Tab.~\ref{table:datasets}. 

Early works focused on rendering composite RGB views~\cite{fisher2012example}. It trains a probabilistic model to generate synthetic data based on hand-created scenes. 
Later, SceneNet~\cite{handa2016scenenet} introduces a generator model to annotate 2D semantics. As depth channel is proved to be helpful for navigation agents~\cite{mirowski2017learning, tai2016towards}, 2D-3D-S~\cite{armeni2017joint} provides assets with depth information. 
Different from these works that render a single room at once, later works~\cite{song2017semantic, deitke2020robothor, kolve2017ai2} provides a large number of scenes consist of bedrooms, living rooms, bathrooms, kitchens, etc. 
However, the synthetic view used by the aforementioned datasets is quite different from the real world scene, which limits the application of the datasets. 
To this end, Matterport3D~\cite{chang2017matterport3d} provides photo-realistic panoramic views by 3D reconstruction and 2D and 3D semantics of these views. 
Gibson~\cite{xia2018gibson} provides a more diverse dataset with 572 houses. 
Replica~\cite{straub2019the} proposes a dataset with 18 indoor scene consist of dense meshes and high-resolution textures. 
Some work such as AI2-THOR~\cite{kolve2017ai2}, RoboTHOR~\cite{deitke2020robothor} and CHALET~\cite{yan2018chalet} rely on the datasets that not currently released.  
The rendering scenes of some datasets are shown in Fig.~\ref{fig:datasets}. 

\subsection{Embodied Simulators}
\label{sec:embodied_simulators}
An embodied simulator provides an interface for an agent to interact with the environment. 
We compare different features of the existing simulators in Tab.~\ref{table:simulator}. 
A simulators is equipped with many sensors, such as a RGB sensor, a depth sensor, a physical sensor and a position sensor.
Early simulators provide low RGB resolution and unrealistic imagery due to the limit of 3D rendering technology.
The lack of visual detail,  limits the navigation performance of the agent. 
Afterwards, to address this, simulations such as Matterport3D simulator~\cite{anderson2018vision}, Gibson simulator~\cite{xia2018gibson} and Habitat~\cite{savva2019habitat} propose high-resolution photo-realistic panoramic view to simulate a more realistic environment.  
Rendering frame rate is also important to embodied simulators since it is critical to training efficiency. 
MINOS~\cite{savva2017minos} runs over 100 frames per second (FPS), which is 10 times faster than its previous works.  
Habitat~\cite{savva2019habitat} over than 1000 FPS on $512 \times 512$ RGB-D image, making it become the fastest simulator among existing simulators. 
Discrete state space in~\cite{anderson2018vision} simplifies the navigation problem and makes the agent easy to learn complex vision-language navigation tasks. 
However, continuous state space is more welcome since it facilitates transferring a learned agent to a real-world robot. 
A customizable simulator is able to generate more diverse data by moving the objects, changing the textures of objects and reconfiguring the lights. 
Diverse data has little bias and therefore, enables the deep learning to learn a robust navigation policy. 
Despite of navigating to find the target object in a static room, interacting is another key skill for real-world robots. 
Some complex tasks may require a robot to interact with objects, such as picking up a cup, moving a chair, or opening a door. 
AI2-THOR~\cite{kolve2017ai2}, iGibson~\cite{xia2020interactive} and RoboTHOR~\cite{deitke2020robothor} provide interactive environments to train such a skill. 
Multi-agent reinforcement learning~\cite{littman1994markov, tan1997multi} is an emerging problem of cooperation and competition among agents.
AI2-THOR and iGibson also support multi-agent training in studying cooperative tasks. 

\begin{table*}
	\centering
	\resizebox{1.0\textwidth}{!}{
	\setlength{\tabcolsep}{0.2em}
    {\renewcommand{\arraystretch}{1.05}
	\begin{tabular}{|l|c|cccc|ccc|}
		\hline
        Simulator & Year & Use Dataset(s) & Resolution & Physics & FPS & Customizable & Interactive & Multi-agent \\ 
		\hline 
		MINOS~\cite{savva2017minos} & 2017 & SUNCG, Matterport3D & $84 \times 84$ & \cmark & 100 & \xmark & \xmark & \xmark \\
		AI2-THOR*~\cite{kolve2017ai2} & 2017 & - & $300 \times 300$ & \cmark & 120 & \cmark & \cmark & \cmark \\
		House3D~\cite{wu2018building} & 2018 & SUNCG & $120 \times 90$ & \xmark & 600 & \cmark & \xmark & \xmark \\
		CHALET~\cite{yan2018chalet} & 2018 & CHALET & $800 \times 600$ & \xmark & 10 & \xmark & \xmark & \xmark \\
		Matterport3D~\cite{anderson2018vision} & 2018 & Matterport3D & $512 \times  512$ & \xmark & 1,000 & \xmark & \xmark & \xmark \\
		Gibson~\cite{xia2018gibson} & 2018 & Gibson,  Matterport3D, 2D-3D-S & $512 \times  512$ & \cmark & 400 & \xmark & \xmark & \xmark \\
		iGibson~\cite{xia2020interactive} & 2018 & Gibson & $512 \times 512$ & \cmark & 400 & \cmark & \cmark & \cmark \\
		Habitat~\cite{savva2019habitat} & 2019 & Matterport3D, Gibson, Replica & $512 \times 512$ & \cmark & 10,000 & \xmark & \xmark & \xmark \\
		RoboTHOR*~\cite{deitke2020robothor} & 2020 & - & $300 \times 300$ & \cmark & 1200 & \cmark & \cmark & \xmark \\
		\hline
	\end{tabular}
	}
	}
	\caption {Comparison of existing embodied simulators (*: the dataset that the simulator uses is not currently released). }
	\label{table:simulator}
	\vspace{-10pt}
\end{table*}

\section{Embodied Navigation Benchmarks}
\label{sec:embodied_navigation_benchmarks}

Here, we introduce several tasks to study the embodied visual navigation problem.
These tasks can be divided into three categories: target-driven navigation task, cross-modal navigation task, and interactive navigation task.

\subsection{Target-driven Navigation Tasks}

\noindent\textbf{PointGoal Navigation}, firstly defined by Anderson \emph{et  al.}~\cite{anderson2018on}, is a task where an agent is initialized to a random starting position and orientation then asked to navigate to a target position. The target position is indicated by its relative coordinates to the starting position. 
This task requires an agent to estimate the cumulative distance from the starting position so that the agent knows how far away from the goal. 
Theoretically, this task is able to be applied to all embodied environments. 

\noindent\textbf{ObjectGoal Navigation} is proposed by Zhu~\emph{et  al.}~\cite{zhu2017target}. In this task, an agent is initialized to a random starting position and is required to find a specific object, such as a desk or a bed. 
Once the navigation agent find the object, it stops. 
The navigation process is regarded as a \emph{success} if the agent is located within a distance to the target object. 
In addition to the room structure, the \emph{ObjectGoal} navigation task needs the object labels and locations. 
Object recognizing and exploring are key skills to the \emph{ObjectGoal} navigation. 

\noindent\textbf{RoomGoal Navigation} is proposed by Wu~\emph{et  al.}~\cite{wu2018building}. In this task, an agent initialized at a random position is asked to navigate to a room (e.g. bedroom or kitchen). 
The navigation process is regarded as a \emph{success} if the agent is stopped within the target room. 
\emph{RoomGoal} navigation requires the room annotations. 
The concept of the room is a high-level semantic. Therefore, a \emph{RoomGoal} navigation agent needs to understand the scene based on the visual details, such as furniture type and room layout. 

\noindent\textbf{Multi-Object Navigation (MultiON)}
Recently, more and more researchers are paying attention to long-term navigation where an agent memorize all the visited scenes. 
Motivated by this, Wani \emph{et al.}~\cite{wani2020multion} propose MultiON, a benchmark for Multi-Object Navigation. 
In MultiON, an agent is asked to navigate to multiple target objects one-by-one, which makes the navigation trajectory quite long. 
The agent raise a FOUND action when it reaches the instructed target. 
Perception and effective planning under partial observation would be the key to solve this task. 

\subsection{Cross-modal Navigation Tasks}

\noindent\textbf{Vision-and-Language Navigation (VLN)} 
VLN is a task where an agent navigates step-by-step following natural language instructions~\cite{anderson2018vision}. 
Previous tasks such as \emph{ObjectGoal} and \emph{RoomGoal} hard-code the object and room semantics as a one-hot vector. On the contrary, VLN introduce natural language sentences to instruct the navigation process like \emph{``Head upstairs and walk past the piano through an
archway directly in front. Turn right when the hallway ends at pictures and table. Wait by the moose antlers hanging on the wall''}. 
The VLN task is successfully completed if the agent stops close to the intended goal following the instruction.

There are several datasets have been proposed for vision-language navigation: R2R~\cite{anderson2018vision}, R4R~\cite{jain2019stay}, and RxR~\cite{ku2020room}. 
The room-to-room (R2R) dataset is proposed in~\cite{anderson2018vision} to study vision-language navigation. 
The R2R dataset contains 21,567 navigation instructions with an average length of 29 words. 
However, the R2R dataset has several shortcomings: 1) the referenced paths are direct-to-goal so that R2R instructions lack the capability of describing complex paths; 2) the instruction consists of several sentences and not fine-grained; 3) the training data is small, and the model is easily overfitting; 4) the language of instruction is English only, and no other languages are included. To address these problems, more advances datasets have been proposed. 
Jain \emph{et al.}~\cite{jain2019stay} 
cross-connects the trajectories and instructions in R2R and generate a new dataset named R4R. 
FGR2R~\cite{hong2020sub} enriches R2R with sub-instructions and their corresponding trajectories. 
RxR~\cite{ku2020room} is a time-aligned dataset and it relieves the
known biases in trajectories and elicits more references to visible entities in R2R. 



\noindent\textbf{Navigation from Dialog History (NDH)}
When navigating in an unfamiliar environment, a human usually asks for assistance and continued navigation according to the  responses of other humans.
However, building an agent that is able to autonomously ask natural language questions and react to the answer is still a long-term goal in robotic navigation. 
In NDH~\cite{thomason2019vision}, an agent is required to navigate according to a dialog history, which consists of several question-answering pairs. Studying NDH is fundamental for building a real-world dialog navigation robot. 

\noindent\textbf{Embodied Questioning and Answering (EQA)}
Visual Question Answering (VQA)~\cite{antol2015vqa} is a cross-modal task, in which a system answers a text-based question with a given image.
VQA soon became one of the most popular computer vision tasks, because it revealed the possibility of interaction between human beings and artificial intelligence agents in natural language~\cite{anderson2018bottom, krishna2017visual, yang2016stacked}.
Compared with VQA, a more advanced activity is to answer questions by self-exploration in an unseen environment. 
Embodied Questioning and Answering (EQA)~\cite{das2018embodied} is a task where an agent is spawned at a random location in a 3D environment and asked a question. 
EQA is a challenging task since it requires a wide range of AI skills: visual perception, language understanding, target-driven navigation, commonsense reasoning, etc. 
In addition to navigation accuracy in other tasks, EQA propose EQA accuracy to measure whether the agent correctly answers the question or not. 

\noindent\textbf{REVERIE}
Recently, Qi \emph{et al.}~\cite{qi2020reverie} proposes Remote Embodied
Visual referring Expression in Real Indoor Environments, named REVERIE in short, to research associating natural language instructions and the visual semantics. 
Different from VLN that gives an instruction that describes the trajectory step-by-step toward the target, the natural language instruction in REVERIE refers to a remote target object.
Compared with \emph{ObjectGoal} navigation, REVERIE offers rich language descriptions to enable the agent to find a unique target in the house. 

\noindent\textbf{Audio-visual Navigation}, proposed by Chen \emph{et al.}~\cite{chen2020soundspaces} introduces audio modality for embodied navigation environment. 
This task requires the agent to navigate to a sound object by seeing and hearing. 
It encourages researchers to study the role of audio plays in navigation. 
This work also offer the SoundSpaces~\cite{chen2020soundspaces} dataset for the 
Audio-visual Navigation task. 
The SoundSpaces dataset is built upon two simulators, Replica and Matterport3D. 
It contains 102 natural sounds across a wide variety of categories: bell, door opening, music, people speaking, telephone, etc. 

\begin{table*}
	\centering
	\resizebox{1.0\textwidth}{!}{
	\setlength{\tabcolsep}{0.3em}
    {\renewcommand{\arraystretch}{1.7}
	\begin{tabular}{|l|cccccccc|}
		\hline
        Metric Name & $\uparrow \downarrow$ & Formulation & PS & SP & UO & SI & OS & CC \\ 
		\hline 
		Path Length (PL) & - & $\sum_{1 \le i < |P|} d(p_i, p_{i+1})$ & - & \cmark & - & \xmark & \xmark & $O(|P|)$ \\
		Navigation Error (NE) & $\downarrow$ & $d(p_{|P|}, r_{|R|})$ & \xmark & \cmark & \xmark & \xmark & \xmark & $O(1)$ \\
		Oracle Navigation Error (ONE) & $\downarrow$ & $\min_{p \in P}d(p, r_{|R|})$ & \xmark & \cmark & \xmark & \xmark & \xmark & $O(|P|)$ \\
		Success Rate (SR) & $\uparrow$ & $\indicator{\text{NE}(P, R) \le d_{th}}$ & \xmark & \xmark & \xmark & \cmark & \xmark & $O(1)$ \\
		Oracle Success Rate (OSR) & $\uparrow$ & $\indicator{\text{ONE}(P, R) \le d_{th}}$ & \xmark & \xmark & \xmark & \cmark & \xmark & $O(|P|)$ \\
		Success weighted by PL (SPL)  &  $\uparrow$   & $\text{SR}(P,R) \cdot \dfrac{d(p_1, r_{|R|})}{\max\{\text{PL}(P), d(p_1, r_{|R|})\}}$ & \xmark & \cmark & \cmark & \cmark & \xmark & $O(|P|)$ \\
		Success weighted by Edit Distance (SED) & $\uparrow$ & $\text{SR}(P,R) \left(1 - \dfrac{\text{ED}(P, R)}{\max{\{|P|, |R|}\} - 1}\right)$ & \cmark & \xmark & \cmark & \cmark & \xmark & $O(|R| \cdot |P|)$ \\
		Path Coverage (PC) & $\uparrow$ & $\dfrac{1}{|R|} \sum_{r \in R}\exp\left(-\dfrac{d(r, P)}{d_{th}}\right)$ & \cmark & \cmark & \cmark & \cmark & \xmark & $O(|R| \cdot |P|)$ \\
		Length Score (LS) & - & $\dfrac{1}{1 + |1 - \frac{\text{PL}(P)}{\text{PC}(P, R) \cdot \text{PL}(R)}|}$ & \xmark & \cmark & \xmark & \xmark & \xmark & $O(|R| \cdot |P|)$ \\
		Coverage weighted by LS (CLS) & $\uparrow$  & $\text{PC}(P, R) \cdot \text{LS}(P, R)$ & \cmark & \cmark & \cmark & \cmark & \xmark & $O(|R| \cdot |P|)$ \\
		Normalized Dynamic Time Warping (nDTW) & $\uparrow$  & $\exp\left({-\dfrac{\min\limits_{W \in \mathcal{W}} \sum_{(i_k, j_k) \in W} d(r_{i_k}, p_{j_k})}{|R|\cdot d_{th}}}\right)$ & \cmark & \cmark & \cmark & \cmark & \cmark & $O(|R| \cdot |P|)$ \\
		\hline
	\end{tabular}
	}
	}
	\caption {We compare the existing metrics from several aspects, including performance ($\uparrow$ indicates the higher the better, $\downarrow$ indicates the lower the better), Formulation, Path similarity (PS), Soft Penalties (SP), Unique Optimum (UO), Scale Invariance (SI), Order Sensitivity (OS), and Computational Complexity (CC). 
	Suppose we have a predicted trajectory $P$ and a ground truth trajectory $R$. 
    $p_i$ and $r_i$ are the ith node on trajectory $P$ and $R$. 
    $|P|$ and $|R|$ stand for the length of $P$ and $R$ respectively. 
	The Dijkstra distance of the house has been preprocessed and the computation complexity of any $d(p_i, r_j)$ is $O(1)$.  }
	\label{table:metric}
	\vspace{-8pt}
\end{table*}

\noindent\textbf{Multi-Target Embodied Questioning and Answering (MT-EQA)}
The natural language questions in EQA are simple since each of them describes one object and lacks attributes and relationships between multiple targets. 
In MT-EQA~\cite{yu2019multi}, the instructions are like \emph{``Is the dresser in the bedroom bigger than the oven in the kitchen"}, where the \emph{dresser} and the \emph{oven} locate in different places with different attributes. 
Thus the agent has to navigate to multiple places, find all targets, analyze the relationships between them, and answer the question. 


\subsection{Interactive Navigation Tasks}

\noindent\textbf{Interactive Questioning and Answering (IQA)}
Building an agent which is able to interact with a dynamic environment is a long-standing goal of the AI community. 
Recently proposed interactive simulators~\cite{xia2020interactive, deitke2020robothor, kolve2017ai2} provide basic functions like opening a door or moving a chair, which enables researchers to build an interactive navigation agent. 
Interactive Questioning and Answering (IQA)~\cite{gordon2018iqa} asks an agent to answer questions by interacting with objects in an environment. 
IQA contains 76,800 training questions that include existence questions, counting questions, spatial relationship questions. 

\noindent\textbf{``Help, Anna!'' (HANNA)}
HANNA~\cite{nguyen2019help} is an object-finding task that allows an agent to request help from Automatic Natural Navigation Assistants (ANNA) when it gets lost. 
Different from NDH that provides a global dialog history as the instruction, the HANNA offers an environment where the instructions dynamically change by the situation. 
The environment creates an interface that enables a human to help the agent when it gets lost in testing time.



\subsection{Evaluation Metrics}
\label{sec:evaluation_metrics}
Many evaluation metrics have been proposed to evaluate how well a navigation agent performs.
We divide them into two categories: trajectory-insensitive metrics and the trajectory-sensitive metrics. 
\noindent\textbf{Trajectory-insensitive Metrics} 
Zhu \emph{et al.}~\cite{zhu2017target} use the average number of steps (i.e., average trajectory length) it takes to reach a target from a random starting point. 
However, there is a large proportion of trajectories fail when the navigation environment becomes more complex and the navigation task becomes more challenging. 
Later works~\cite{wu2018building, gupta2017cognitive, savva2017minos} introduce propose the Success Rate (SR) to measure frequency of the agent successfully reach the goal and other works~\cite{gupta2017cognitive, das2018embodied} report Navigation Error (NE), the mean distance toward the goal when the agent finally stops. 
Oracle Success Rate (OSR) is proposed to evaluate if the agent correctly stops following the oracle stopping rule~\cite{misra2017mapping, anderson2018vision}. 
These metrics measure the probability of whether the agent completes the task or not, however, fail to measure how much proportion it completes the task. 

\noindent\textbf{Trajectory-sensitive Metrics} 
Success weighted by Path Length (SPL) is the first metric that evaluates both the efficiency and efficacy of a navigation agent, and it is regarded as the primary metric in VLN. 
The SPL ignores the turning actions and the agent heading. 
\emph{Success weighted by edit distance} (SED)~\cite{chen2019touchdown} takes turning actions into consideration and fix this problem. 
SED is designed for instruction compliance in a graph-based environment, where there exists a certain correct path. 
However, in some tasks like R4R~\cite{jain2019stay} and R6R~\cite{zhu2020babywalk}, the instructed paths are not direct-to-goal. 
Therefore, it is not appropriate to evaluate the navigation performance of the SPL. 
Therefore, \emph{Coverage weighted by Length Score} (CLS)~\cite{jain2019stay} is proposed to measure the fidelity of the agent's behavior to the described path. 
CLS is the product of two variables: path coverage and length fraction. 
Ilharco~\emph{et al.} absorb the idea of Dynamic Time Warpping~\cite{berndt1994using}, an approach widely used in various areas~\cite{sakoe1978dynamic, vakanski2012trajectory, keogh2000scaling}, and propose normalized Dynamic Time Warping (nDTW) metric~\cite{ilharco2019general} to evaluate the navigation performance. 
Similar to CLS, nDTW evaluates the distance between the predicted path with the ground-truth path. 
Moreover, nDTW is sensitive to the order of the navigation path while CLS is order-invariant. 
nDTW can be implemented in an efficient dynamic programming algorithm. 
The path-sensitive metrics, like CLS and nDTW, perform better when they are used as reward functions than target-oriented reward functions in reinforcement learning to navigate~\cite{jain2019stay, ilharco2019general}. 

\noindent\textbf{Measurements of Metrics} 
Each metric has its unique characteristics according to their formulation. 
We compare the formulation and characteristics of existing metrics in Tab.~\ref{table:metric}. 
In this part, we introduce measurements to evaluate the functions of a metric: 

\noindent1) \textbf{Path Similarity (PS)} characterizes a notion of similarity between the $P$ and the $R$. This implies that metrics should depend on all nodes in $P$ and all nodes in $R$. PS penalizes deviations from the ground truth path, even if they lead to the same goal. This is not only prudent, as agents might wander around undesired terrain if this is not enforced, but also explicitly gauges the fidelity of the predictions with respect to the provided language instructions. 

\noindent2) \textbf{Soft Penalties (SP)} penalizes differences from the ground truth path according to a soft notion of dissimilarity that depends on distances in the graph. This ensures that larger discrepancies are penalized more severely than smaller ones and that SP should not rely only on dichotomous views of intersection. 

\noindent3) \textbf{Unique Optimum (UO)} yields a perfect score if and only if the reference and predicted paths are an exact match. This ensures that the perfect score is unambiguous: the reference path $R$ is therefore treated as a golden standard. No other path should have the same or higher score as the reference path itself. 

\noindent4) \textbf{Scale Invariance (SI)} measures if a metric is independent over different datasets. If a metric variants over datasets, such as navigation error, its scores across different datasets cannot be directly compared.  

\noindent5) \textbf{Order Sensitive (OS)} indicates if a metric is sensitive to the navigation order with the same trajectory length, success rate, etc. The navigation order reveals some sorts of navigation policy even though it is usually hard to be evaluated. 

\noindent6) \textbf{Computational Complexity (CC)} measures the cost of computing a pair of $(P, R)$. It is important to design a fast algorithm to calculate the score for automatic validation and testing. 

\subsection{Summary}
Embodied navigation benchmarks define the tasks and metrics for different settings. 
The target-oriented tasks like \emph{PointGoal}, \emph{ObjectGoal}, and \emph{RoomGoal} Navigation can provide label by the 3D assets and do not require extra human annotation. 
cross-modal navigation tasks like R2R~\cite{anderson2018vision}, Visual Dialogue Navigation~\cite{thomason2019vision} or REVERIE~\cite{qi2020reverie} require human to label the trajectory and the corresponding language description. 
The interactive interactive tasks~\cite{nguyen2019help, xia2020interactive} require the agent to learn to manipulate objects, which attract rising attention due to their wide application in real-world scenarios. 

\section{Methods in Simulated Environments}
\label{sec:methods_simulated_env}

In this section, we mainly discuss two problems in the simulated environments: target-driven navigation and cross-modal navigation. And we introduce the methods to solve these problems. 

\subsection{Target-driven Navigation}
\label{sec:target_navigation}
Methods for this problem focus on navigating from a random starting position to a target. 
The target may be specified by an RGB image, a vector, or a word. 
The agent predict actions like \emph{turn left}, \emph{turn right}, \emph{move forward} to navigate in the embodied environment and predict \emph{stop} indicate the stop action. There are diverse methods that try to solve this problem, including: 1) model-free methods; 2) planning-based methods; and 3) self-supervised methods. 

\subsubsection{Model-free Methods}

\begin{figure}[t]
	\centering
	\includegraphics[width=0.99\linewidth]{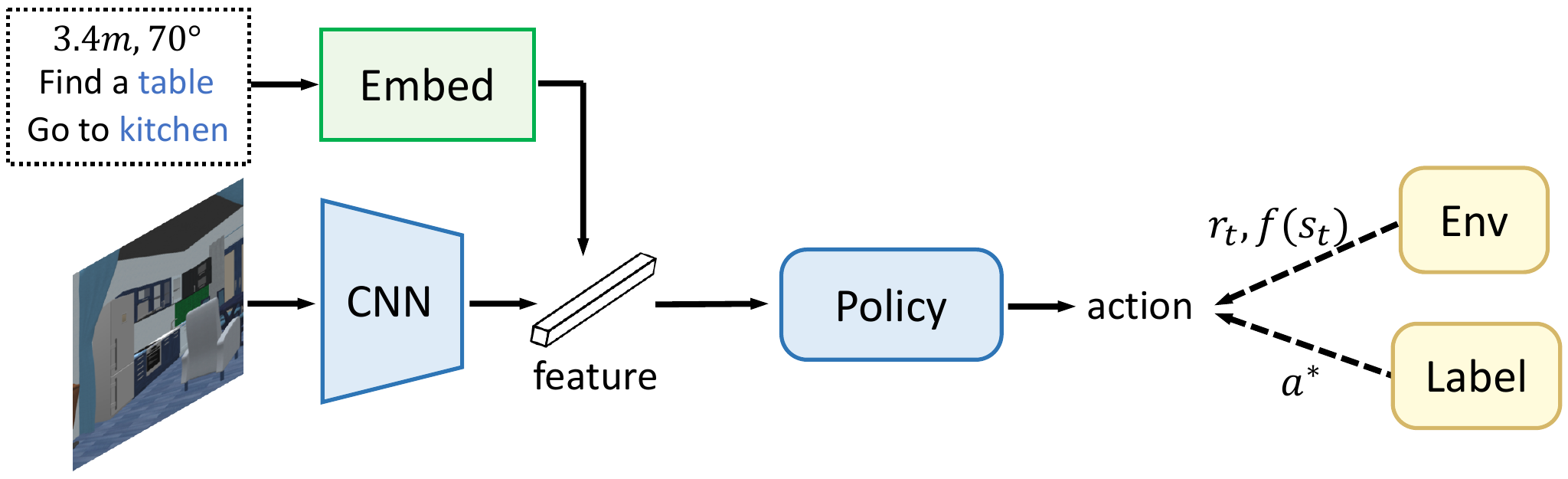}
	\caption{
	An illustration of an model-free visual navigation model. This model learned from imitation learning and reinforcement learning. $r_t$ is the reward and $f(s_t)$ stands for the labels calculated from the state $s_t$. And $a'$ is the label stands for the optimal action. }
	\label{fig:model-free}
	\vspace{-8pt}
\end{figure}

The model-free methods learn to navigate end-to-end without modeling the environment, as illustrated in the Fig.~\ref{fig:model-free}. 
The learning objective includes imitation learning or reinforcement learning. 
The formulation of the learning object is: 
\begin{equation}
\mathcal{L} = \sum_t -a_t^*log\left(p_t\right)-\sum_t a_t log\left(p_t\right) A_t, 
\end{equation}
where $a^*$ is the ground truth action, $p_t$ is the action probability, and $A_t$ is the advantage in A3C~\cite{mnih2016asynchronous}. 
Though extensive reinforcement learning works~\cite{tamar2017value, lee2018gated, misra2017mapping} have long studied 2D navigation problem where an agent receives global state for each step, the embodied navigation problem with partial observation remains challenging. 
Many robot control works~\cite{lenser2003visual, royer2005outdoor, haddad1998reactive, remazeilles2004robot} focus on obstacle avoidance rather than trajectory planning. 

Zhu \emph{et al.}~\cite{zhu2017target} firstly propose to use deep learning for feature matching and deep reinforcement learning for policy prediction, which allows the agent to better generalize to unseen environments. 
Afterwards, Successor Representation (SR)~\cite{zhu2017visual} is proposed to enable the agent to interact with objects. 
This framework takes the states of objects and a discrete description of the scene into consideration. 
Successor Representation encodes semantic information and concatenate it with the visual representation as in~\cite{zhu2017target}. 
Different from~\cite{zhu2017target} that only uses reinforcement learning to learn a policy predictor, Successor Representation model that bootstraps reinforcement learning with imitation learning. 
Previous models lack of the ability of encoding temporal information. 
By introducing an LSTM layer to encode historical information, Wu \emph{et al.}~\cite{wu2018building} are able to build an agent that is able to generalize to unseen scenarios.
In the ablation study, this work proves that A3C~\cite{mnih2016asynchronous} outperforms DDPG~\cite{lillicrap2016continuous} in visual navigation task, 
and the model learned from semantic mask outperforms which learned from RGB inputs. 
Inspite of solving visual navigation problem via on-policy deep reinforcement learning algorithms, some works adopt other algorithms. 
Li \emph{et al.}~\cite{li2020learning} propose an end-to-end model based on Q-learning that learns viewpoint invariant and target on invariant visual servoing for local mobile robot navigation. 

There are lots of works use segmentation masks of objects to augment visual inputs. 
Mousavian \emph{et al.}~\cite{mousavian2019visual}
exploit the instance features in the vision inputs by introducing Faster-RCNN detector trained on MSCOCO dataset~\cite{lin2014microsoft} and a segmenter defined by~\cite{mousavian2016joint} to detect and segment objects. 
Shen \emph{et al.}~\cite{shen2019situational} improve zero-shot
generalization of a navigation agent by fusing diverse visual representations, including RGB features, depth features, segmentation features, detection features, etc. 
The different visual representations are adaptively weighted for fusing. 
To further improve the robustness, they propose a inter-task affinity regularization that encourages the agent to select more complementary and less redundant representations to fuse. 
%
Despite the well-performed detector and segmenter, 
learning a robust navigation policy is still challenging. 
For example, to search for mugs, a human would search cabinets near the coffee machine and for fruits a human may try the fridge first. 
To address this, Lv \emph{et al.}~\cite{lv2020improving} integrate 3D knowledge graph and sub-targets into deep reinforcement learning framework. 
To enhance the cross-scene generalization, Wu \emph{et al.}~\cite{wu2021reinforcement} introduce an information theoretic regularization term into the RL objective and models the action-observation dynamics by learning a variational generative model. 


Some works investigate problem settings other than indoor navigation, such as street view navigation or combining other modalities. 
Khosla \emph{et al.}~\cite{khosla2014looking} firstly attempt 
to solve outdoor street navigation task by embodied visual navigation method ,where the agent navigate purely based on panoramic street views. 
DeepNav~\cite{brahmbhatt2017deepnav} is build upon a Convolutional Neural Network
(CNN) for navigating in large cities using locally visible street-view images. 
These works rely on supervised training with the ground truth compass input , however, the compass can sometimes be unavailable in real-world. 
Another work~\cite{mirowski2018learning} propose an end-to-end deep reinforcement learning framework that uses the street scenes from Google Street View as visual input but without the ground truth compass. Recognizing the importance of locale-specific knowledge to navigation, they propose a dual pathway architecture that allows locale-specific features to be encapsulated. 
AV-WaN~\cite{chen2021learning} is proposed to tackle the challenges in Audio-visual Navigation.  
This model learns the audio-visual waypoints and dynamically sets intermediate goal locations based on its audio-visual observations and partial maps in an end-to-end manner. 

\begin{figure}[t]
	\centering
	\includegraphics[width=0.99\linewidth]{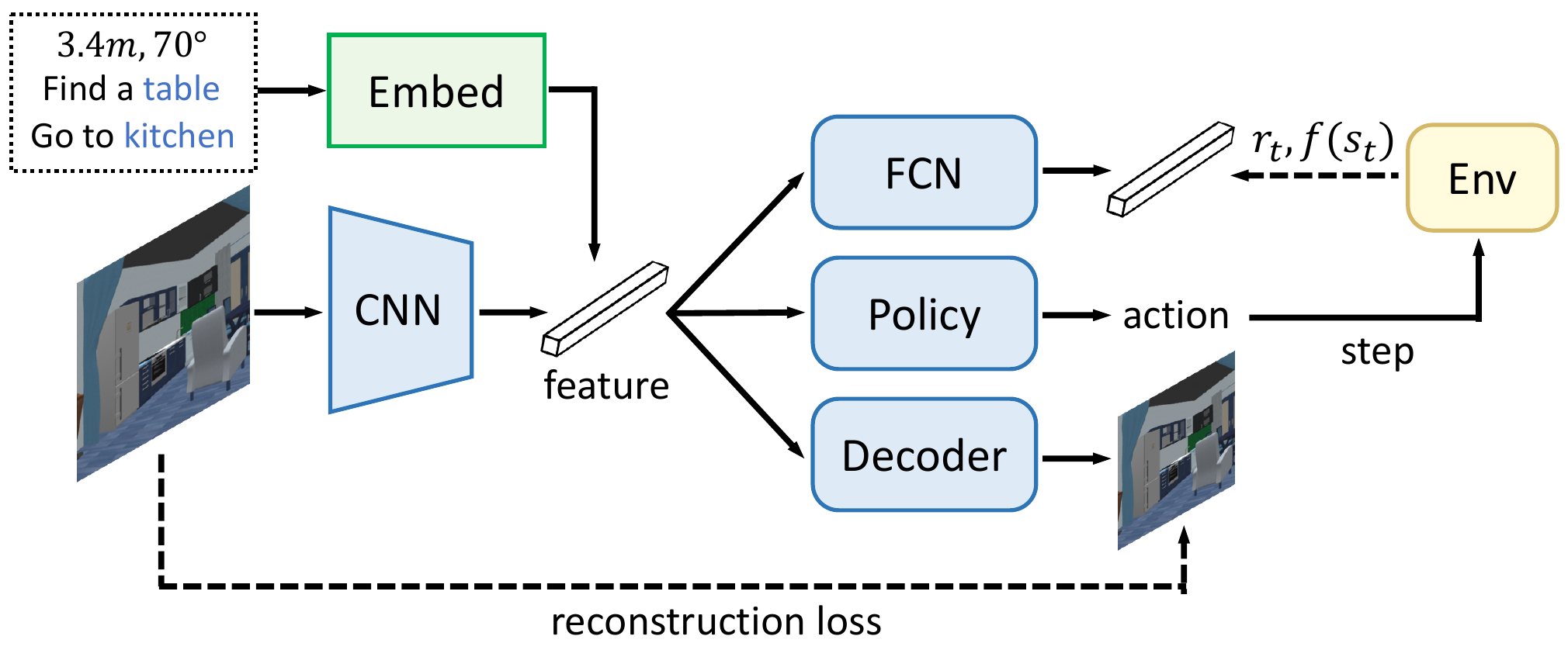}
	\caption{
	An illustration of an end-to-end visual navigation model with self-supervised objectives. $r_t$ is the reward and $f(s_t)$ stands for the labels calculated from the state $s_t$. }
	\label{fig:self-sup}
	\vspace{-8pt}
\end{figure}

\subsubsection{Self-Supervised Methods} 

Self-supervised learning is a long studied topic of exploiting extra training signals via various pretext tasks. 
It enables an agent to learn more knowledge without additional human annotations. 
Various self-supervised tasks have been proposed in the field of deep learning, such as context prediction~\cite{doersch2015unsupervised}, solving jigsaw puzzles~\cite{noroozi2016unsupervised}, colorization~\cite{zhang2016colorful}, rotation~\cite{sun2020test}. 
There is also some auxiliary tasks proposed to improve data efficiency and generalization in reinforcement learning. 
Xie \emph{et al.}~\cite{xie2016model} combines self-supervised learning with model-based reinforcement learning to solve robotic tasks. 
Motivated by traditional UVFA architecture~\cite{schaul2015universal} which learns a value function by means of feature learning, Jaderberg \emph{et al.}~\cite{jaderberg2017reinforcement} invent auxiliary control and reward prediction tasks that dramatically improve both data efficiency and robustness. 

In embodied navigation, the environment contains unstructured semantic information that is hard to learn in end-to-end manner. 
In spite of explicitly modeling the environment using SLAM or memory mechanism, self-supervised learning provide another feasible way of learning the unstructured knowledge. Mirowski~\emph{et al.}~\cite{mirowski2017learning} propose an online navigation model with two self-supervised auxiliary objectives, predicting the current depth view by RGB view and detecting the loop closure. 
%
Similiar idea~\cite{dosovitskiy2016learning} has been applied in game applications~\cite{kempka2016vizdoom} for rapid exploration. 
%
Auxiliary tasks also can speed up learning. 
Ye \emph{et al.}~\cite{ye2020auxiliary, ye2021auxiliary} achieve great success in \emph{PointGoal} and \emph{ObjectGoal} navigation by assembling reinforcement learning with various kinds of auxiliary tasks, formulated as: 
\begin{equation}
    L_{total}=L_{RL}+\sum_i^n \beta_i L_{Aux,i}. 
\end{equation}
Visual perception is critical for visual navigation. 
But the training signal provided by reinforcement learning contain too much noise to train a robust feature perception network. 
An encoder-decoder architecture is proved to be beneficial~\cite{ye2019gaple} in visual encoding and segmentation predicting. In addition, an auxiliary task is used to penalize the segmentation error, which benefits the learning of feature perception. 
However, this self-supervised auxiliary task only learns the low-level dynamic function between two adjacent states and fails to learn the high-level semantic information. 
To guarantee the semantic consistency of actions in a trajectory, Liu \emph{et al.}~\cite{liu2017learning} propose an auxiliary regularization task to penalizes the inconsistency of representations. 
This regularization task encourages the policy network to extract salient features from each sensor. 
Real-world robot locomotion is far from deterministic due to the presence of actuation noise, which might be caused by wheels slipping, motion sensor error, rebound, etc. 
To reduce the noise, Datta \emph{et al.}~\cite{datta2020integrating} introduce an auxiliary task of localization estimation by means of temporal difference. 
The auxiliary task is used to train a CNN network and use the estimated locomotion as an input of the policy network. 
A curiosity-driven self-supervised objective~\cite{bigazzi2021explore} is applied to encourage exploration while penalizing the repeating actions. A stable curiosity-driven policy without repeating actions could improve the exploration efficiency. 
Self-supervised auxiliary tasks are also helpful in cross-modal understanding for navigation. 
Dean \emph{et al.}~\cite{dean2020see} use audio as an additional modality for self-supervised exploration. 
It includes an curiosity driven intrinsic reward, which encourages the agent to explore novel associations between different sensory modalities (audio and visual). 
An overview of the pipeline of self-supervised navigation methods is shown in Fig.~\ref{fig:self-sup}. 
The agent firstly embeds a visual image and an instruction as features. 
Then the visual feature and the instruction feature are fused to predict the action. The auxiliary tasks use the fused feature to make a prediction, such as predicting the reward, or reconstructing the input visual image. 

\begin{figure}[t]
	\centering
	\includegraphics[width=0.85\linewidth]{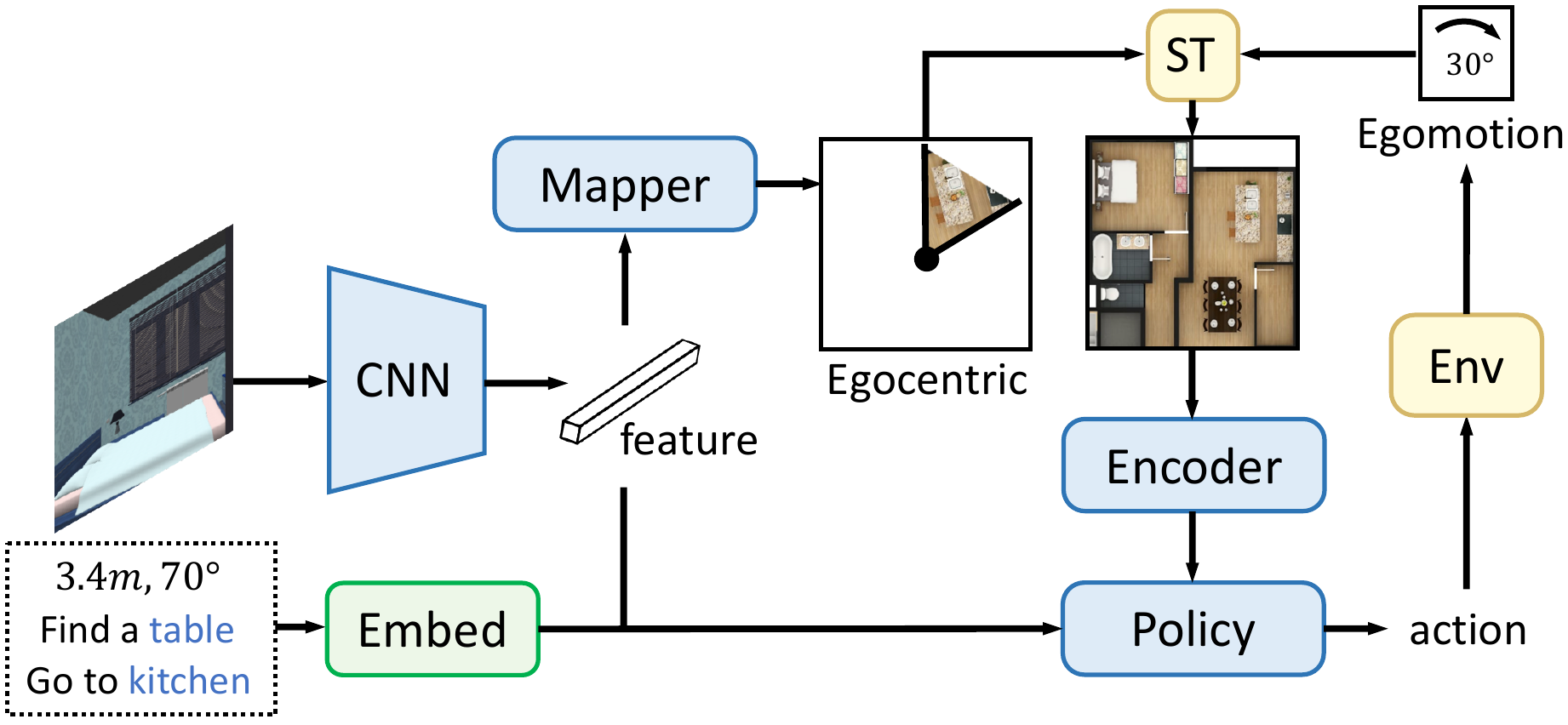}
	\vspace{4pt}
	\caption{
	An overview of the common practice of the ``Neural SLAM''-based model. ``ST'' is the spatial transformation function.  }
	\label{fig:slam}
	\vspace{-8pt}
\end{figure}

\subsubsection{Planning-based Methods} 

The map building problem for an unknown environment while solving the localization problem at the same
time is known as Simultaneous Localization and Mapping (SLAM)~\cite{durrant2006simultaneous, filipenko2018comparison}. 
The earlier investigations on visual navigation
were carried out with a stereo camera~\cite{se2002mobile, olson2003rover} and a
monocular camera, such as  MonoSLAM~\cite{davison2003real}. 
Over the past decade, traditional geometric-based approaches~\cite{engel2017direct, engel2014lsd, mur2017orb} remains dominating the field. 
With the development of deep learning, some methods like CNN-SLAM~\cite{tateno2017cnn}, DVO~\cite{wang2018learning} and D3VO~\cite{yang2020d3vo} are proposed. 
Some indoor tasks are proposed to study SLAM, such as KITTI~\cite{geiger2012we} and EuRoC~\cite{burri2016euroc}. 
However, these tasks are different from embodied navigation task. 
The odometry benchmark is to estimate the location given a sequence of visual inputs while the navigation task is to align the instruction with the environment semantics. 

Recently, researchers find that the ability of localization is important to navigation, especially for long-term path planning. 
Thus, some works introduce SLAM methods to model the house and improve the localization ability of the agent.
Neural Map~\cite{parisotto2017neural} generalize this idea for all deep reinforcement learning agents rather than navigation only. 
However, this work assumes the location of the agent is always known and does not utilize the 2D structure of this memory. 
Neural SLAM~\cite{zhang2017neural} fixes this problem by embedding SLAM-like procedures into the soft-attention~\cite{vaswani2017attention}. 
To avoid spatial blurring associated with repeated warping, MapNet~\cite{henriques2018mapnet} proposes to use a world-centric rather than an egocentric map. 
Different from previous works, MapNet maintains a 2.5D representation by a deep neural network module that learns to distill visual representations from 3D embodied visual inputs. 
Gordon \emph{et al.}~\cite{gordon2018iqa} proposes Hierarchical Interactive Memory Network (HIMN), a framework with hierarchical controller for IQA task. 
The high-level controller is a planner that decides the long-term navigation target, and the low-level controller predicts the action, interacts with the environment, and answers the question. 
Gupta \emph{et al.}~\cite{gupta2020cognitive} introduce the Neural-SLAM method in embodied navigation. This work consists of two parts: mapping and planning. 
The mapping mechanism maintains a 2D memory map. For each step, it transforms the embodied scene into a 2D feature and update the map with the feature. 
The planning mechanism uses a value function to output a policy. 


Efficient exploration is widely regarded as one of the main challenges in reinforcement learning (RL)~\cite{jaksch2010near, jin2018is, azar2017minimax}. 
Similarly, it is important in navigation since the target does not always visible from the starting position and the agent is required to explore the unseen scene and search for the target. 
Recently, exploration based on explicitly modeled semantic memory is proven to be efficient. 
To learn a policy with spatial memory, Chen \emph{et al.}~\cite{chen2019learning} bootstrap the model with imitation learning and finetune it with coverage rewards derived purely from on-board sensors. 
Active Neural SLAM (ANS)~\cite{chaplot2020learning} is a successful neural SLAM method which achieves the state-of-the-art on the CVPR 2019 Habitat \emph{Pointgoal} Navigation Challenge. 
ANS proposes a hierarchical structure for planning. 
Inspired by the idea of hierarchical RL~\cite{barto2003recent, dayan1992feudal}, ANS learns the high-level planner by reinforcement learning and learns the low-level planner by imitation learning. 
The mapper is implemented by an auxiliary task of predicting a 2D map. 
The first channel of the map stands for if there is an obstacle and the second map stands for whether the position has been explored or not. 
However, the predefined 2D map cannot help long-term navigation since the semantic information of the scenes in different viewpoints is not encoded in the map. 
Neural Topological SLAM~\cite{chaplot2020neural} propose a more advanced way which stores the observed feature representations. 
This method introduce a graph update module to leverage semantics. 
The graph update module maintains a topological feature memory. 
For each step, the module localize current observation into memory nodes. 
If an observation is not localized in any node of the memory, the graph update module will add a new node into the topological feature memory. 
Goal-Oriented Semantic Exploration (SemExp)~\cite{chaplot2020object} tackles the object goal navigation task in realistic environments. 
This method first builds a episodic semantic map and uses it
to explore the environment based on the category of the target object. 
This approach achieves state-of-the-art in Habitat ObjectNav Challenge 2020. 
An overview of the common practice of the `Neural SLAM'-based model is shown in Fig.~\ref{fig:slam}. 
In addition to the visual encoder and the instruction encoder as in Fig.~\ref{fig:self-sup}, 
`Neural SLAM'-based model have a unique module to project a embodied visual view to feature representation and store it in a 2D top-down map: 
\begin{equation}
    m_t, \hat{x_t}=f_{SLAM}(s_t, x_{t-1:t}', m_{t-1}|\theta_{S})
\end{equation}
where $x_{t-1:t}'$ stands for previous poses, $m_{t-1}$ is previous maps, and $\theta_S$ stands for parameters. 
This map models the room structure and visual representation of scenes. The projected feature representations are fused with the visual feature and the instruction feature to jointly predict an action. 


\subsubsection{Summary} 

Compared with traditional robotics methods, the model-free methods are able to obtain robust navigation models by sampling large scale data with the embodied simulator. 
Some works adopt detection and segmentation approaches to get better visual views. 
In spite of indoor scenarios, model-free methods achieve great success in street scene and multi-modal environments. 
Self-supervised methods are proposed to exploit the extra knowledge by auxiliary tasks to improve the learning efficiency and generalization ability. 
Planning-based methods utilize a 2D map or a topological memory to model the environment during navigation. 

\subsection{Cross-modal Navigation}
\label{sec:navigation_with_natural_language}
A navigation robot who understands natural language can accomplish more complex tasks, such as ``\emph{pick up the cup in the kitchen}" or ``\emph{help me find my glass upstairs}". 
In this section, we introduce three kinds of works that solves cross-modal navigation tasks: 
1) step-by-step methods; 2) pretraining-based methods; 3) planning based methods. 

\subsubsection{Sequence-to-sequence Navigation}


Anderson \emph{et al.}~\cite{anderson2018vision} firstly propose a sequence-to-sequence model similar to~\cite{sutskever2014sequence} to address the vision language navigation problem. 
This model sequentially encodes a language instruction word-by-word, concatenates the sentence feature with the visual image feature and decodes the action sequence. 
However, a sequence-to-sequence model is lack of stability and generalization since it fails to consider the dynamics in the real-world environments. 
RPA~\cite{wang2018look} is proposed to tackle the generalization issue by equipping a `look-ahead'
module, which learns to predict the future state and reward. 
To improve the generalization ability in instruction-trajectory alignment, Fried \emph{et al.}~\cite{fried2018speaker} propose a data augmentation approach named ``speaker-follower'' to improve the model generalization. 
To generate augmentation data, the speaker firstly translates a randomly trajectory into an instruction, and the follower secondly translates an instruction into a trajectory: 
\begin{equation}
\underset{r \in R(d)}{\mathrm{argmax}} P_S(d|r) \cdot P_F(r|d)^{(1-\gamma)}, 
\end{equation}
where $P_S$ is the speaker, $P_F$ is the follower, $d$ stands for an instruction, $r$ stands for a trajectory, and $\gamma$ is a weighting factor. 
Another contribution of this paper~\cite{fried2018speaker} is the definition of a high-level action that move forward toward an orientation in a panoramic space in stead of low-level actions like \emph{turn left}, \emph{turn right} and \emph{go forward}.  
Compared with the definition of low-level actions, this approach largely reduce the length of the action sequence that describes the same trajectory. 
Navigating by the high-level action space requires less prediction times, which makes the model easier to train and more robust to test. 
Howeverr, previous methods learn to navigate by imitation learning only with the instruction-trajectory data pairs, which supervises the shortest path while ignore the sub-optimal trajectories so that leds to overfitting. 
To tackle this problem, Wang \emph{et al.}~\cite{wang2018reinforced} propose to jointly learn a navigation agent by imitation learning and reinforcement learning. 
\begin{figure}[t]
	\centering
	\includegraphics[width=0.95\linewidth]{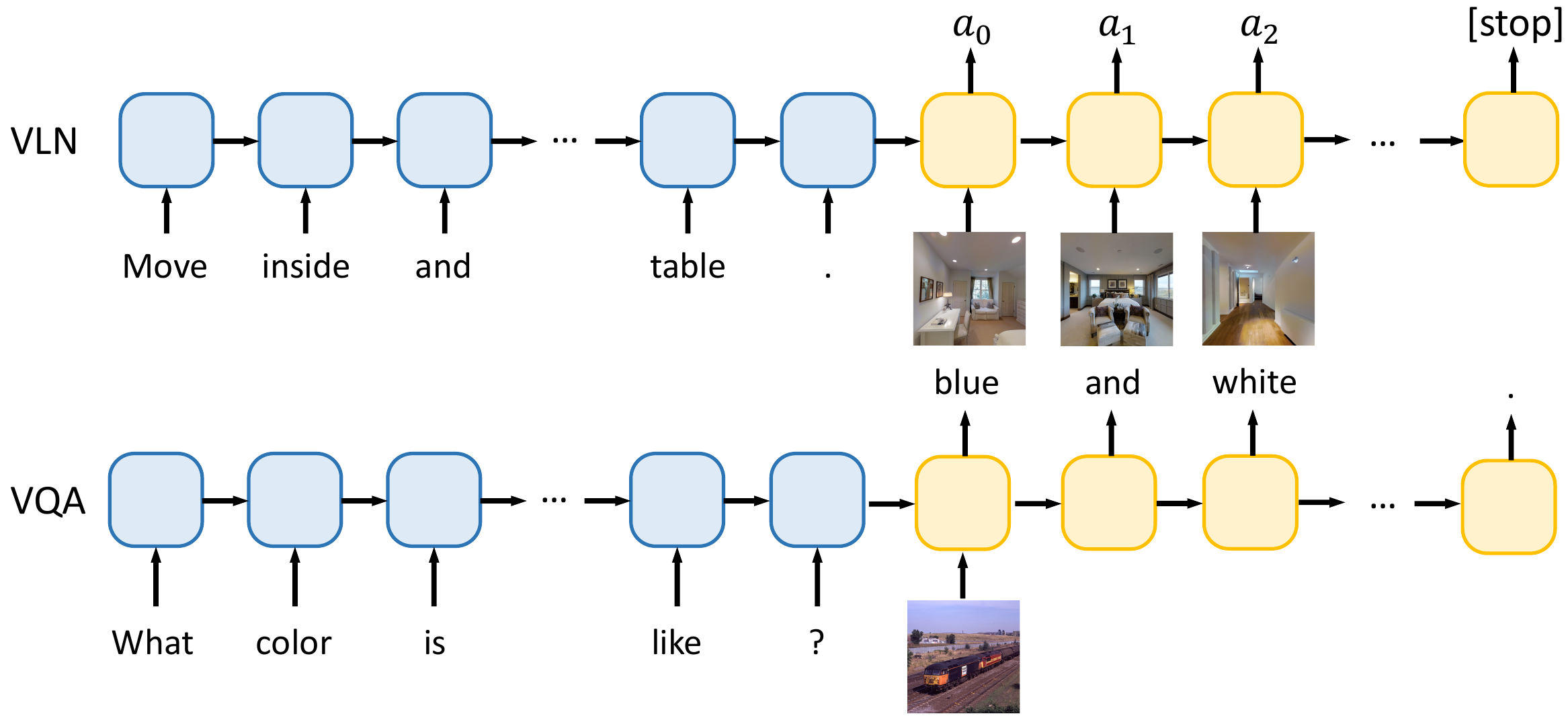}
	\vspace{-5pt}
	\caption{
	A comparison of seq-to-seq models in VLN and VQA. }
	\label{fig:seq-seq}
	\vspace{-8pt}
\end{figure}
In addition, this method introduce an LSTM to encode the temporal information of visual features and introduce a cross-modal mechanism to achieve better vision-language navigation ability. 
Ma \emph{et al.}~\cite{ma2019self} propose a self-monitoring agent with a visual-textual co-grounding module and progress monitor. 
The progress monitor use the cross-modal feature from the co-grounding module and estimate the completed progress. 
Since the instruction in vision-language task guides the agent to the target step-by-step, the progress information contain rich knowledge that help improve the perception of the agent. 
Ma \emph{et al.} propose in~\cite{ma2019the} the \emph{Regretful Agent}, with a regretful module which uses the estimated progress to indicate if the agent navigates to a wrong place and need to go back. 
Similar to the \emph{Regretful Agent}, Ke \emph{et al.} propose ~\cite{ke2019tactical} 
a framework for using asynchronous search to boost a VLN navigator by enabling explicit backtrack. 
Anderson \emph{et al.}~\cite{anderson2019chasing} regard the step-by-step navigation process as a visual tracking task. This approach implements the navigation agent within the framework of Bayesian state tracking~\cite{thrun2005probabilistic} and formulates an end-to-end differentiable histogram filter~\cite{jonschkowski2016end} with learnable observation and motion models. 
One commonly used method that relieve the visual overfitting is to apply an dropout~\cite{srivastava2014dropout} layer on the visual feature, which is extracted by a pretrained network like VGG~\cite{simonyan2015very} or ResNet~\cite{he2016deep}. 
Tan \emph{et al.}~\cite{tan2019learning} argue that simply applying a dropout layer on the visual feature leads to inconsistency, e.g. a chair in this frame could be dropped in the next frame. 
To solve the problem, They propose a environmental dropout layer that randomly dropout some fixed channels during a trajectory. 
Zhu \emph{et al.}~\cite{zhu2020vision} propose AuxRN, a framework that introduce self-supervised auxiliary tasks to exploit environmental knowledge from several aspects. 
In addition to introducing the temporal difference auxiliary task that is widely use in other embodied visual navigation methods~\cite{ye2021auxiliary, mirowski2017learning}, AuxRN introduces a trajectory retelling task and instruction-trajectory matching task that learn the temporal semantics of a trajectory. 
Instead of generating the low-quality augmented data, Fu \emph{et al.}~\cite{fu2019counterfactual} introduce the concept of counterfactual thinking to sample challenging paths to augment training dataset. 
They present a model-agnostic adversarial path sampler (APS) to pick the difficult trajectories and only consider useful counterfactual conditions. 

Different from the earlier works that based on data augmentation and other classical navigation methods, some works discover the importance of natural language to VLN. 
Thomason \emph{et al.}~\cite{thomason2019shifting} find the unimodal baseline outperforms random baselines and even
some of their multimodal counterparts. 
Thus the work advocates that ablating unimodal to evaluate the bias is important to proposing a dataset. 
A study of Huang \emph{et al.}~\cite{huang2019multi} shows that only a limited number of those augmented paths in ~\cite{fried2018speaker}
are useful and after using 60\% of the augmented data, the improvement diminishes with additional augmented data. 
To avoid the extensive work in reward engineering, Wang \emph{et al.}~\cite{wang2020soft} propose a Soft Expert Reward Learning model that includes two parts: 1) soft expert distillation, which encourages agents to behave like an expert in soft fashion; 2) self perceiving, which pushes the agent towards the final destination as fast as possible. 
Xia \emph{et al.}~\cite{xia2020multi} leverages multiple instructions as different descriptions for the same trajectory to resolve language ambiguity and improve generalization ability. This work indicates that the human annotations in VLN are largely biased according to the specific scene and the trajectory. 
The quality of visual features is critical for improving the performance of embodied navigation. 
Previous works extract global visual features from panoramic views by a pretrained CNN network like ResNet-101~\cite{he2016deep}. 
Hong~\emph{et al.}~\cite{hong2020language} introduce Faster-RCNN to detect objects in navigation and build a relationship graph between visual and language entities for vision-language alignment. 
In spite of the visual inputs, the structure information also helps navigation. 
Hu \emph{et al.}~\cite{hu2019are} discover that the language instructions contain high-level semantic information while visual representations are a lower-level modality, which makes the vision-language alignment difficult. 
Motivated by this, they decomposes the grounding procedure into a set of expert models with access to different modalities and ensemble them at prediction time. 
To better research what role does language understanding play in VLN task, 
Hong \emph{et al.}~\cite{hong2020sub} argue that the intermediate supervision is important in vision-language alignment. Thus, they propose FGR2R, 
a method which enables navigation processes to be traceable and encourage the agent to move at the level of sub-instructions. 

\begin{figure}[t]
	\centering
	\includegraphics[width=0.99\linewidth]{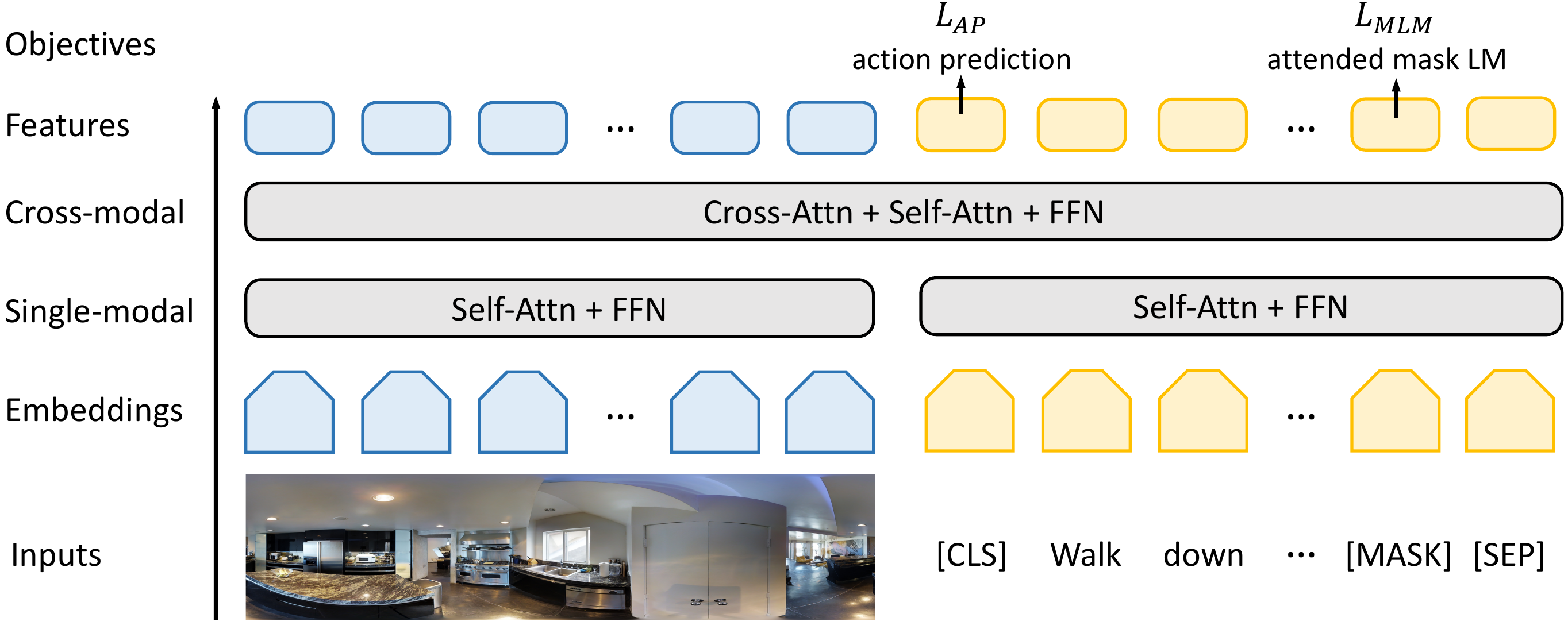}
	\vspace{4pt}
	\caption{
	An example of pretraining-based framework. }
	\label{fig:pretrain}
	\vspace{-8pt}
\end{figure}

\subsubsection{Pretraining-based Methods} 

Several challenges are discovered during the research on the vision-language navigation: 1) low training efficiency; 2) large data bias (include both vision and language); 3) lack of generalization from seen to unseen scenes. 
To address these challenges, pretraining-based models are proposed to learn from large-scale data sets from other sources and fast adapt to unseen scenarios.

\noindent\textbf{Low training efficiency} The traditional encoder-decoder framework first samples the total trajectory by teacher-forcing or student-forcing and then back-propagate the gradients. 
In other deep learning tasks like image classification~\cite{krizhevsky2017imagenet} or text recognition~\cite{shi2016robust}, the model predicts a result directly. 
However, in the vision-language navigation task, the agent predicts a trajectory by interacting with the environment in a step-by-step manner, which is so time-consuming that reduce the training efficiency. 

\begin{figure*}[t]
	\centering
	\includegraphics[width=0.9\linewidth]{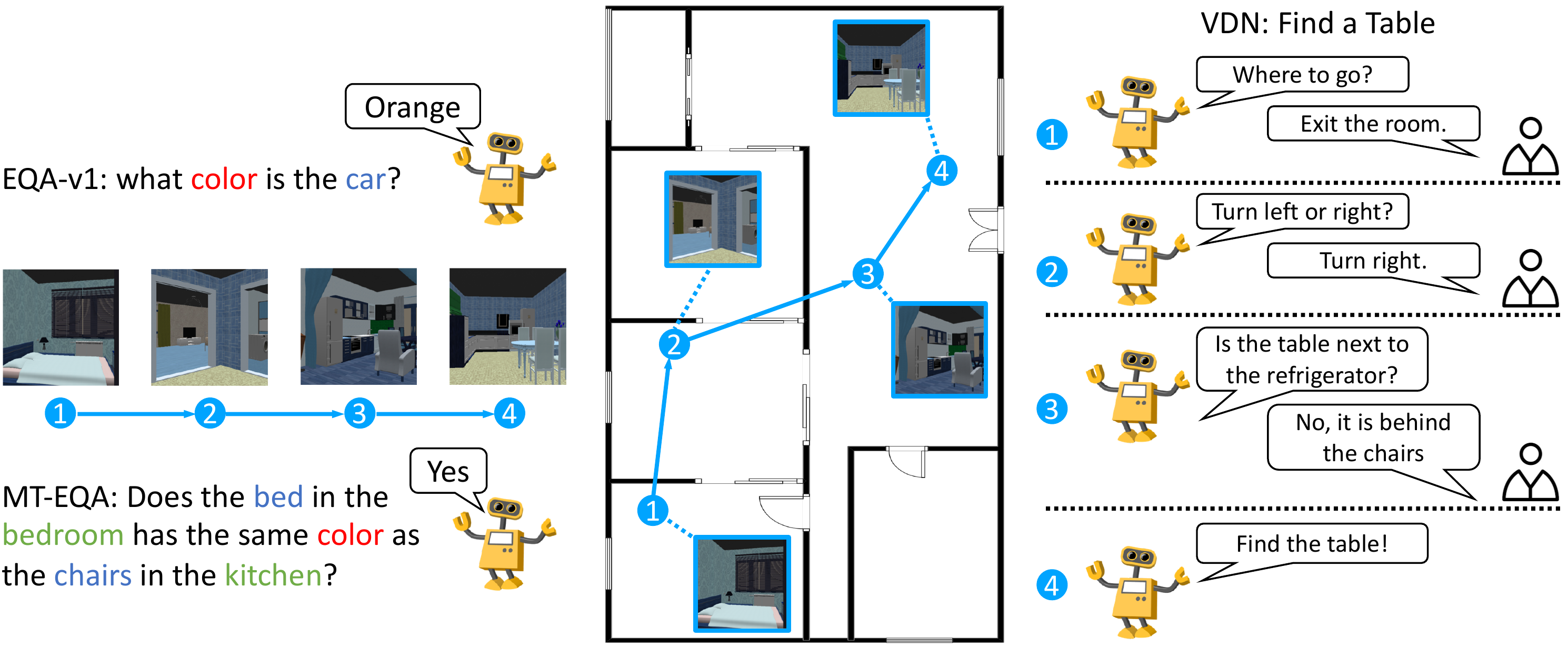}
	\vspace{-5pt}
	\caption{
	A comparison of three embodied vision-language navigation tasks: EQA-v1~\cite{das2018embodied}, MT-EQA~\cite{yu2019multi} and VDN~\cite{thomason2019vision}. }
	\label{fig:eqa_vdn}
	\vspace{-8pt}
\end{figure*}

\noindent\textbf{Large data bias} 
The vision-language navigation scenarios are so diverse that 61 houses in R2R cannot cover all of them. 
From the aspect of natural language, in the R2R task, only 69\% of bigrams are shared between training and evaluation. 

\noindent\textbf{Lack of generalization} 
Lacking of diverse training data still largely limits the generalization in spite of the proposed augmentation methods like trajectory augmentation, visual feature augmentation and natural language augmentation. 
Thus, introducing extra knowledge from other tasks and datasets becomes a promising topic.  

Pretraining-based methods largely improve  the generalization ability of a model by learning in large scale of data~\cite{he2016deep, anderson2018bottom}. 
Furthermore, bert-based methods~\cite{devlin2018bert, tan2019lxmert} pretrain a transformer network with proxy tasks and achieve great success in vision, language and cross-modal tasks. 
Many researchers consider to solve the vision-language navigation problem by pretaining-based methods. 
Li \emph{et al.}~\cite{li2019robust} propose \textsc{PreSS} first introduce a pretrained language models to learn instruction representations. And they propose a stochastic sampling scheme
to reduce the gap between the expert actions in training and the sampled actions
in testing. 
Majumdar \emph{et al.}~\cite{majumdar2020improving} advocate to improve model by leveraging large-scale of web data. 
However, it is hard to transfer the static image data to VLN task. 
Therefore, they propose VLN-bert, a transformer-based model which is pretrained by static images and their captions. 
\textsc{Prevalent}~\cite{hao2020towards} is self-supervisedly pretrained on large amount of image-text-action triplets sampled from an embodied environment with two pretrianing objectives, masked language modeling (MLM) and action prediction (AP): 
\begin{align}
\begin{split}
    &L_{MLM}=-\mathbb{E}_{s \sim p(\tau), (\tau,x) \sim D_{E}} 
    \mathrm{log}\ p(x_i|\textbf{x}, s), \\
     &L_{AP}=-\mathbb{E}_{(a,s) \sim p(\tau), (\tau,x) \sim D_{E}} 
    \mathrm{log}\ p(\textbf{x}|x_{[CLS]}, s), 
\end{split}
\end{align}
where $(s, a)$ a state-action pair. 
\textsc{Prevalent} is proven to be effective on several vision-language navigation datasets, including R2R, CVDN and HANNA. 
The embodied navigation agent receives partial observation rather than global observation, which is better to be modeled as a partially observable Markov Decision Process. 
Different from the encoder-decoder model, previous pretraining-based models do not memorize previously seen scenes during navigation and utilize temporal knowledge, which causes information loss in action prediction. 
Motivated by this, Hong \emph{et al.}~\cite{hong2021vln} propose a recurrent multi-layer transformer network that is time-aware for use in VLN. This method introduce a Transformer which maintains a feature vector to represent temporal context.

\subsubsection{Navigation with Questioning and Answering} 

Instead of passively perceive natural language instructions from a human commander, Das \emph{et al.}~\cite{yu2019multi} suggest that an intelligent agent should be able to answer a question via navigation. 
Thus Das \emph{et al.} present a new task named EQA (Embodied Questioning and Answering), where an agent is spawned at a random location in a 3D environment and asked to answer a question. 
In order to answer, the agent have to first navigate to explore the environment, gather information through egocentric vision, and then answer the question. 
To solve this challenging task, Das \emph{et al.} present PACMAN, a CNN-RNN model with Adaptive Computation Time (ACT) module~\cite{graves2016adaptive} to decide how many times to repeatly execute an action~\cite{yu2019multi}. 
The PACMAN is bootstrapped by shortest path demonstrations and then fine-tuned with RL. 
However, this method is lack of the ability of high-level representation. 
In a later work~\cite{eqa_modular}, Das \emph{et al.} propose a hierarchical policy named Neural Modular Controller (NMC) that operates at multiple
timescales, where the higher-level master policy proposes sub-goals to be executed
by low-level sub-policies. 
Anand \emph{et al.}~\cite{anand2018blindfold} find that a blindfold (question-only) baseline on EQA and find that the baseline perform previous state-of-the-art models. 
They suggest that previous EQA models are ineffective at leveraging the context from the environment and the EQAv1 dataset has lots of noise. 

Wu \emph{et al.}~\cite{wu2020revisiting} propose a simple supervised learning baseline which is competitive to the state-of-the-art EQA methods. 
To improve EQA performance in unseen environment, in this paper, they propose a setting in which allows the agent to answer questions for adaptation. 
Yu \emph{et al.}~\cite{yu2019multi} argues that the EQA task assumes that each question has exactly one target, which limits its application. 
Therefore, Yu \emph{et al.} present Multi-Target EQA (MT-EQA), a generalized version of EQA. The question of this task contains multiple targets. And it require the agent to perform comparative reasoning over multiple targets rather than simply perceive the attributes of one target. 
Wijmans \emph{et al.}~\cite{eqa_matterport} extend the EQA problem to photorealstic environment. 
In this environment, they discover that point cloud representations are more effective for navigation. 
Luo \emph{et al.}~\cite{luo2019segeqa} suggest that the visual perception ability limits the performance of the EQA. They introduce Flownet2~\cite{ilg2017flownet}, a high-speed video segmentation framework as a backbone to assist navigation and question answering. 
Li \emph{et al.}~\cite{li2019walking} propose a MIND module that model the environment imagery and generate mental images that are treated as short-term sub-goals. 
Tan \emph{et al.}~\cite{tan2020multi} investigate the questioning and answering problems between multiple targets. 
In this task, the agent has to navigate to multiple places, find all targets, analysis the relationships between them, and answer the question. 
Motivated by recent progress in Visual Question Answering (VQA)~\cite{antol2015vqa} and Video Question Answering (VideoQA)~\cite{ye2017video},
Cangea \emph{et al.}~\cite{cangea2019videonavqa} propose VideoNavQA, a dataset that contains pairs of questions and videos generated in the House3D environment. 
This dataset fills the gap between the VQA and the EQA. 
The VideoNavQA task represents an alternative view of the EQA paradigm: 
By providing nearly-optimal trajectories to the agent, the navigation problem is easier to solve compared with the reasoning problem.  
Deng \emph{et al.}~\cite{deng2021mqa} propose Manipulation Question Answering (MQA) where the robot is required
to find the answer to the question by actively exploring the environment via manipulation. 
To suggest a promising direction of solving MQA, they provide a framework which consists of a QA module (VQA framework) and a manipulation model (Q learning framework). 
Nilsson \emph{et al.}~\cite{nilsson2020embodied} build an agent which explores in a 3D environment and occasionally requests annotation during navigation. 
Similarly, Roman \emph{et al.}~\cite{roman2020rmm} suggest a two-agent paradigm for cooperative vision-and-dialogue navigation. 
Their model learns multiple-skills, including navigation, question asking, and questioning-answering components. 

\subsubsection{Navigation with Dialogue}
There is a long history that human use a dialog to guide a robot~\cite{thomason2019improving, tellex2014asking}. 
In the field of embodied navigation, Banerjee \emph{et al.}~\cite{nguyen2019help} propose ``Help, Anna!" (HANNA), an interactive photo-realistic simulator in which an agent fulfills object-finding tasks by requesting and interpreting natural language and vision assistance. 
Nguyen \emph{et al.}~\cite{nguyen2019vision} propose a task named VLNA, where an agent is guided via language to find objects. 
However, the language instruction in these two tasks far from real-world problem: the responses of HANNA are automatic generated from a trained model while the guidance of VLNA are in the form of templated language that encodes gold-standard planner action. 
Vries \emph{et al.}~\cite{vries2018talk} propose ``Talk The Walk" (TtW), where two humans communicate to reach a goal location in an outdoor environment. 
However, in TtW, the human uses an abstracted semantic map rather than an egocentric view of the environment. 

Thomason \emph{et al.}~\cite{thomason2019vision} propose vision-and-dialog navigation (VDN), a scaffold for navigation-centered question asking and question
answering tasks where an agent navigates following a multi-round dialog history rather than an instruction. 
Compared with the single-round instructions in R2R dataset, VDN provides multi-round annotation, in which each round of dialogue describes a sub-trajectory. 
The more fine-grained dialogue annotation facilitate researchers to study the problem of navigation with natural language.  
Zhu \emph{et al.}~\cite{zhu2020vision} propose a framework with a cross-modal memory mechanism to capture the hierarchical correlation between the dialogue rounds and the sub-trajectories. 
More generally, several methods, such as \textsc{Prevalent}~\cite{hao2020towards} and \textsc{BabyWalk}~\cite{zhu2020babywalk}, validate their navigation ability using both sentence instructions and dialog instructions. 
Unfortunately, these works heavily rely on dialogue annotations which is labor-intensive. 
To alleviate this, Roman \emph{et al.}~\cite{roman2020rmm} exploit to generate dialogue questions answers based on visual views. 
This work addresses four challenges in modeling turn-based dialogues, which includes: 1) deciding when to ask a question; 2) generating navigator questions; 3) generating question-answer pairs for guidance; 4) generating navigator actions. 
To achieve this, Roman \emph{et al.}~\cite{roman2020rmm} introduce a two-agent paradigm, where one agent navigates and asks questions while the other guides agent answers. 
Different from previous works that guide navigator with template language, this work initialize the oracle model via pretraining on CVDN dialogues to generate natural language. 

A dialog does not always describe a step-by-step navigation process. 
Rather, the oracle describes the target scene and let the navigator to find it, which commonly occurs when someone get lost in a new building. 
Hahn \emph{et al.}~\cite{hahn2020where} propose a LED task (localizing the observer
from dialog history) to realize when it get lost. 
Motivated by this, they present a dataset named \textsc{Where Are You}~\cite{hahn2020where} that consists of 6k dialogues of two humans. 
Due to the wide application of multi-agent communication systems~\cite{jain2019two, singh2018learning} in real-world, researchers become interested in implementing dialog navigating in physical environments. 
Marge \emph{et al.}~\cite{marge2019a} present MRDwH, a platform that implements autonomous dialogue management and navigation of two simulated robots in a large outdoor simulated environment. 
Banerjee \emph{et al.}~\cite{banerjee2020the} propose RobotSlang benchmark, a dataset which is gathered by pairing a human ``driver'' controlling a physical robot and asking questions of a human ``commander''

We compare the difference of Embodied Question Answering (EQA)~\cite{das2018embodied}, Multi-Target Embodied Question Answering (MT-EQA)~\cite{yu2019multi} and Vision-and-dialog navigation (VDN)~\cite{thomason2019vision} in Fig.~\ref{fig:eqa_vdn}. 
We demonstrate three different dialogues for the same navigation trajectory as an example. 
Compared with EQA, the question in MT-EQA are more complex since it should describe multiple targets. 
The agent have to acquire high-level skills, such as reasoning, comparison and multi-object localization, to accomplish MT-EQA. 
In the EQA and MT-EQA tasks, the agent is required to answer question from a human via navigation. 
However, in the VDN task, the agent is the navigator and the questioner which asks a human for hints to find the target. The difference of the task setting led to the different designs of the navigation model. 

\subsubsection{Summary} 
Natural language provides an interface for a human to interact with a robot. 
A robot with cross-modal understanding is able to accomplish complex tasks such as navigating following a natural language instruction or a dialogue, asking the oracle for more details, etc. 
Lots of works have been proposed to research on vision-language navigation problem from diverse aspects. 

\begin{figure*}[t]
	\centering
	\includegraphics[width=0.99\linewidth]{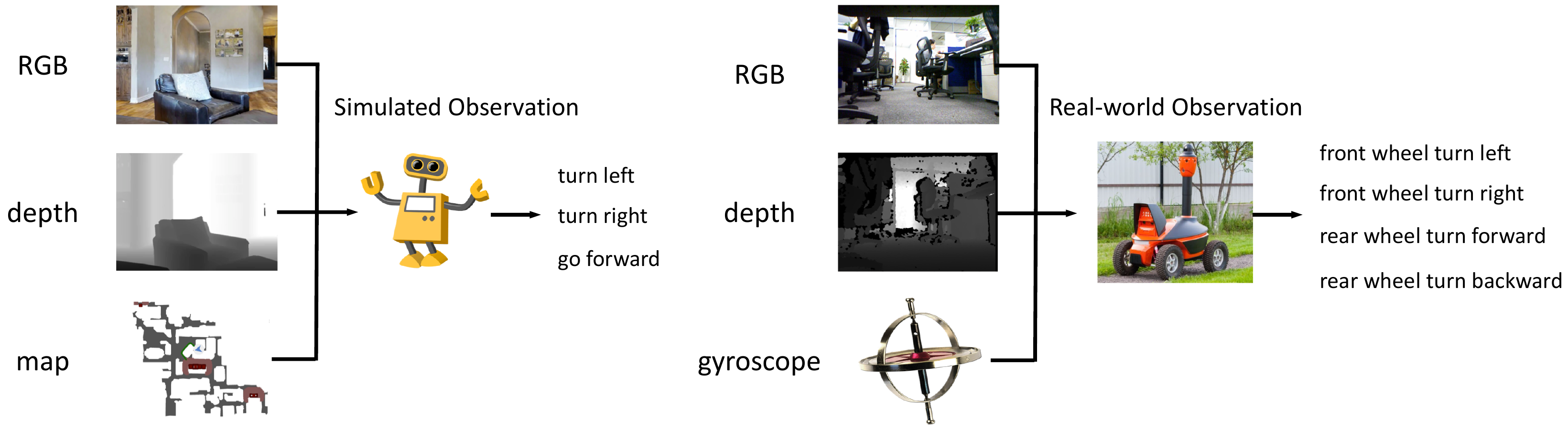}
	\caption{
	A comparison of input space and action space of a simulated environment and the real-world environment. }
	\label{fig:sim_real_env}
	\vspace{-8pt}
\end{figure*}

\section{Methods in Real-world Environments}
\label{sec:real_world_embodied_navigation}
Embodied navigation methods in simulated environments give a promising direction of solving real-world navigation problems. 
In this section, we are going to 1) introduce methods for real-world applications; 2) compare them with the methods in simulators; 3) discuss the possibility of sim-to-real transferring. 

\subsection{Real-world Navigation Methods} 
\label{sec:real_world_navigation_methods}
\subsubsection{Indoor Robotic Navigation} 
Deep learning plays an important role in indoor navigation for real-world applications.
LeCun \emph{et al.}~\cite{muller2006off} firstly adopt convolutional network for obstacle avoidance. 
Hadsell \emph{et al.}~\cite{hadsell2009learning} propose a self-supervised learning process that accurately classifies long-range vision semantics via a hierarchical deep model. 
The method is validated on a Learning applied to ground robots (LAGR)~\cite{jackel2006darpa}. 
Later, more and more real-world robots adopt deep learning to perceive and extract distinctive visual features~\cite{zhang2016deep}. 
Zhang \emph{et al.}~\cite{zhang2017deep} research on the problem where a real robot navigates in simple maze-like environments. 
Based on the success of RL algorithms for solving challenging control tasks~\cite{mnih2015human, lillicrap2016continuous}, Zhang \emph{et al.} employ successor representation in learning to achieve quick adaptation. 
Morad \emph{et al.}~\cite{morad2021embodied} present an indoor object-driven navigation method named NavACL that uses automatic curriculum learning and is easily generalized to new environments and targets. 
Kahn \emph{et al.}~\cite{kahn2018composable} adopt multitask learning and off-policy RL learning to learn directly from real-world events. 
This method enables a robot to learn autonomously and be easily deployed on multiple real-world tasks without any human provided labels. 

\subsubsection{Outdoor Robotic Navigation} 
There has been a long history that human study outdoor navigation robot. 
Thorpe \emph{et al.}~\cite{thorpe1988vision} present two algorithms, a RGB-based method for road following and a 3D-based method for obstacle detection, for a robot to learn to navigate in a campus. 
Ross \emph{et al.}~\cite{ross2013learning} combine deep learning and reinforcement learning to learn obstacle avoidance for UAVs. 
Morad \emph{et al.} evaluate the performance of NavACL on two simulated environments, Gibson and Habitat. And we transfer the navigation to a Turtlebot3
wheeled robot (AGV) and a DJI Tello quadrotor (UAV). 
Both quantitative and qualitative results reveal that the policy of NavACL trained in the simualted environment is surprisingly effective in AGV and UAV. 
Manderson \emph{et al.}~\cite{manderson2020vision} use conditional imitation learning to train an underwater vehicle to navigate close to sparse geographic waypoints without any prior map. 


\subsubsection{Long-range Navigation} 
Their model achieves  the best performance and shows competitive generalization ability on a real robot platform. 
Borenstein \emph{et al.}~\cite{borenstein1990real} propose to maintain a world model~\cite{ha2018world} that updated continuously and in real-time to avoid obstacles. 
The world model learns and simulates the real-world environment and reduce the cost of data sampling~\cite{ha2018world}. 
Liu \emph{et al.} propose Lifelong Federated Reinforcement Learning (LFRL), a learning architecture for navigation in cloud robotic systems to address this problem. 


Long-range navigation is challenging for real-world robots. 
To address this proble, Francis \emph{et al.}~\cite{francis2020long} present PRM-RL, a hierarchical robot navigation method. 
The PRM-RL model consists of a reinforcement learning agent that learns short-range obstacle avoidance from noisy sensors, and a sampling-based planner to map the navigation space. 
Shah \emph{et al.}~\cite{shah2020ving} propose ViNG, a learning-based navigation system for
reaching visually indicated goals and demonstrating this system on a real mobile robot platform. 
Unlike prior work, ViNG uses purely offline experience and does not require a simulator or online data collection, which significantly improves the training efficiency. 
Mapping~\cite{thrun2002probabilistic} and path planning~\cite{lavalle2006planning} has also been widely adopted by many real-world applications. 
Davison \emph{et al.}~\cite{davison1998mobile} builds an automatic system, which is able to detect, store and track suitable landmark features during goal-directed navigation. 
They show how a robot can use active vision to provide continuous and accurate global positioning, thus achieving efficient navigation. 
Sim \emph{et al.}~\cite{sim2006autonomous} enable a robot to accurately localize its location by employing a hybrid map representation of 3D point landmarks.

\section{Navigation from Simulator to Real-world}
\label{sec:challenge_solution_real_navigation}
In this section, we first demonstrate the challenges in the real-world navigation by comparing the difference between simulated environments and the real-world environment. Then we introduce the methods that focus on solving these challenges. 

\subsection{Comparison of Simulated and Real Navigation}
\label{sec:sim-real}
Today, current achievements in the simulated navigation are still far of building a real-world navigation robot. 
Compared with the simulated environments, the real-world navigation environment is much more complex and ever changing. 
An comparison of inputs between a simulated environment (Habitat~\cite{savva2019habitat}) and the real-world environment is shown in Fig.~\ref{fig:sim_real_env}.

\subsubsection{Reasons of Domain Gap} 
We summarize three aspects cause the sim-real domain gap: 1) observation space; 2) action space; 3) environmental dynamics. 

\noindent\textbf{Observation Difference.}
An observation of the simulated environment can be an RGB image, a depth image, or a ground truth map. 
The quality of the RGB image and depth image inputs are high. 
The environment contains all static object information and enable it to provide ground truth information, like room structure, segmentation or object labels. 
The simulated environments provide unreal synthetic images with fewer objects where the real-world environments are far more complex with many.
The sensors in the real-world environment, including RGB, GPS, and the velocity sensor, are usually noisy while the sensors in the simulated environment have no noise. 
Although some simulators~\cite{savva2017minos, savva2019habitat, xia2018gibson} provide physical sensors and simulate some physical interactions (such as collision and acceleration), the performance of their physics engine is still far from real. 

\noindent\textbf{Action Difference.}
Different from the simple action space consists of `turn left', `turn right', and `go forward' in the simulated environment, the action space in the real world is more challenging, depending on the structure of the robot. 
Lots of obstacles exist during real-world navigation, which blocks the robot from turning or moving forward.
Real-world environments are often dynamic since the environment is so complex that many factors are changing in the long term or short term, such as temperature, moisture, friction, obstacles, and pedestrians. 
Another challenge which is also widely ignored in the simulated environments is the complexity and instability of the action space. 
For example, the results of executing the same action are uncertain since the physical condition is evolving, such as the wheels are skidding or get stuck. 

\noindent\textbf{Environmental Dynamics.}
The evolving of environmental conditions, such as temperature, humidity or parts wear cause the environmental dynamics. A policy without online adaptation ability cannot handle this problem well. 
Recently, more and more attention has been paid to the adaptive policy of learning dynamic environment. 
Some works~\cite{berseth2021smirl, nagabandi2018learning, lee2020context} propose simulated robot environments to accomplish this, however, the simulation is far simpler than the real-world. 

\begin{figure}[t]
	\centering
	\includegraphics[width=0.95\linewidth]{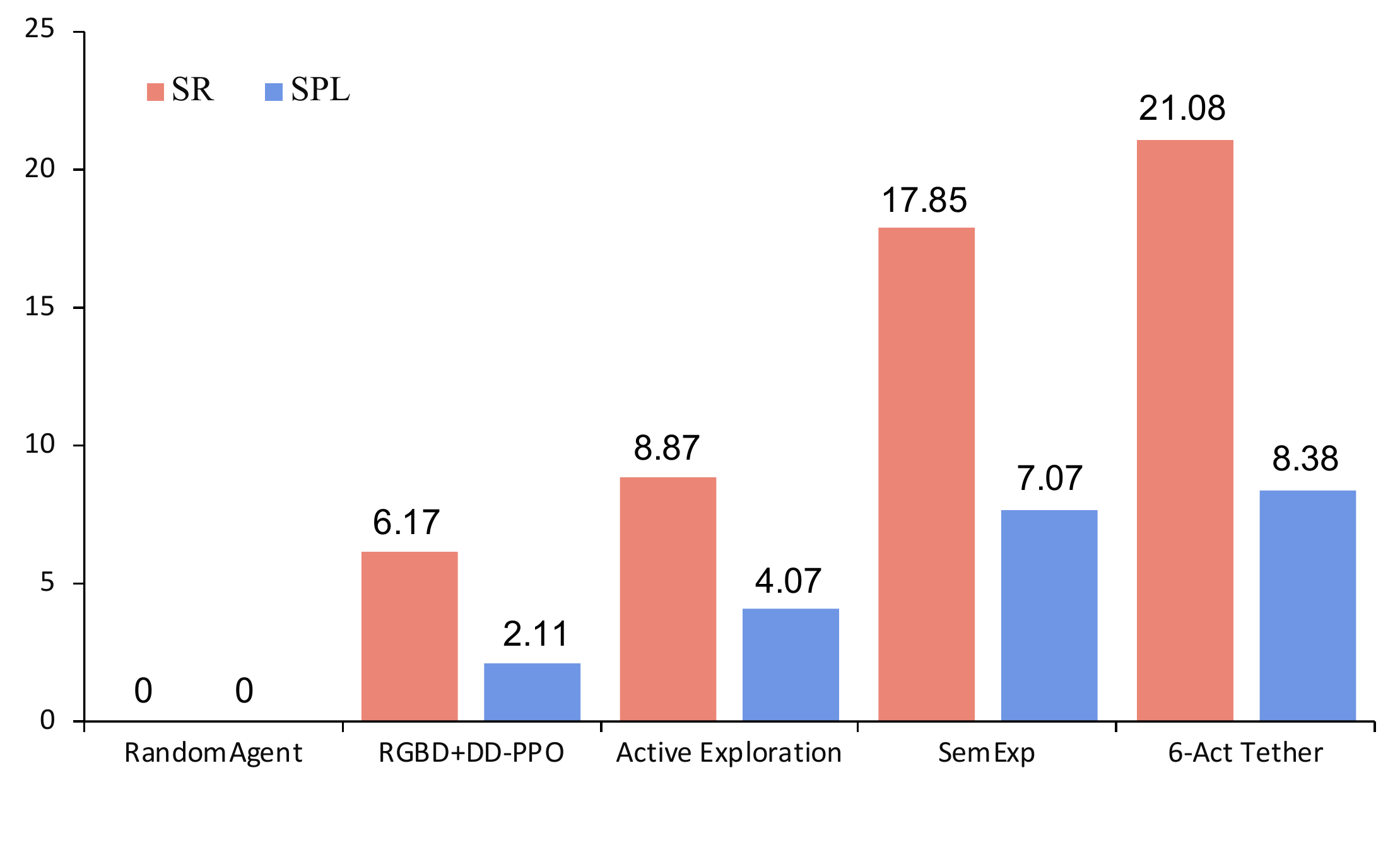}
	\vspace{-8pt}
	\caption{
	The performances of methods on the Habitat \emph{ObjectGoal} navigation, including DD-PPO~\cite{wijmans2020dd}, Active Exploration~\cite{chaplot2020learning}, SemExp~\cite{chaplot2020object} and 6-Act Tether~\cite{ye2021auxiliary}. }
	\label{fig:habitat_o}
	\vspace{-10pt}
\end{figure}

\begin{table*}
	\centering
	\resizebox{1.\textwidth}{!}{
	\setlength{\tabcolsep}{0.6em}
    {\renewcommand{\arraystretch}{1.15}
	\begin{tabular}{|l|cccc|cccc|cccc|}
	\hline
    \multicolumn{1}{|c}{\multirow{2}{*}{Methods}} & \multicolumn{4}{|c}{R2R Validation Seen} & \multicolumn{4}{|c}{R2R Validation Unseen} & \multicolumn{4}{|c|}{R2R Test Unseen} \\
    \cline{2-13} &
    \multicolumn{1}{c}{TL} & \multicolumn{1}{c}{NE$\downarrow$} & \multicolumn{1}{c}{SR$\uparrow$} & \multicolumn{1}{c|}{SPL$\uparrow$} & \multicolumn{1}{c}{TL} & \multicolumn{1}{c}{NE$\downarrow$} & \multicolumn{1}{c}{SR$\uparrow$} & \multicolumn{1}{c|}{SPL$\uparrow$} & \multicolumn{1}{c}{TL} & \multicolumn{1}{c}{NE$\downarrow$} & \multicolumn{1}{c}{SR$\uparrow$} & \multicolumn{1}{c|}{SPL$\uparrow$} \\
    \hline
    Random     & 9.58  & 9.45 & 16   & -    & 9.77  & 9.23 & 16   & -    & 9.89  & 9.79 & 13 & 12 \\
    Human      & -     & -    & -    & -    & -     & -    & -    & -    & 11.85 & 1.61 & 86 & 76 \\
    \hline
    Seq2Seq \cite{anderson2018vision}
        & 11.33 & 6.01 & 39 & -    & 8.39  & 7.81 & 22 & -    & 8.13  & 7.85 & 20 & 18 \\
    Speaker-Follower \cite{fried2018speaker}
        & -     & 3.36 & 66 & -    & -     & 6.62 & 35 & -    & 14.82 & 6.62 & 35 & 28 \\
    RPA \cite{wang2018look}
        & 8.46 & 5.56 & 43 & - & 7.22 & 7.65 & 25 & - & 9.15 & 7.53 & 25 & - \\
    SMNA \cite{ma2019self}
        & -     & 3.22 & 67 & 58 & -     & 5.52 & 45 & 32 & 18.04 & 5.67 & 48 & 35 \\
    RCM+SIL \cite{wang2018reinforced}
        & 10.65 & 3.53 & 67 & -    & 11.46 & 6.09 & 43 & -    & 11.97 & 6.12 & 43 & 38 \\
    Regretful \cite{ma2019the}
        & - & 3.23 & 69 & 63 & - & 5.32 & 50 & 41 & 13.69 & 5.69 & 48 & 40 \\
    PRESS*~\cite{li2019robust}
        & 10.57 & 4.39 & 58 & 55 & 10.36 & 5.28 & 49 & 45 & 10.77 & 5.49 & 49 & 45 \\
    FAST-Short \cite{ke2019tactical}
        & -     & -    & -    & -    & 21.17 & 4.97 & 56 & 43 & 22.08 & 5.14 & 54 & 41 \\
    EnvDrop \cite{tan2019learning}
        & 11.00 & 3.99 & 62 & 59 & 10.70 & 5.22 & 52 & 48 & 11.66 & 5.23 & 51 & 47 \\
    AuxRN \cite{zhu2020vision}
        & -     & 3.33 & 70 & 67 & -     & 5.28 & 55 & 50 & -     & 5.15 & 55 & 51 \\
    PREVALENT*~\cite{hao2020towards}
        & 10.32 & 3.67 & 69 & 65 & 10.19 & 4.71 & 58 & 53 & 10.51 & 5.30 & 54 & 51 \\
    Active Exploration~\cite{wang2020active} 
        & 19.70 & 3.20 & 70 & 52 &  20.60 & 4.36 & 58 & 40 & 21.6 & 4.33 & 60 & 41 \\
    RelGraph \cite{hong2020language} 
        & 10.13 & 3.47 & 67 & 65 & 9.99 & 4.73 & 57 & 53 & 10.29 & 4.75 & 55 & 52 \\
    VLN$\protect\CircleArrowright$BERT*~\cite{hong2021vln}
        & 11.13 & 2.90 & 72 & 68 & 12.01 & 3.93 & 63 & 57 & 12.35 & 4.09 & 63 & 57 \\
    \hline
	\end{tabular}
	}
	}
	\caption {Comparison of agent performance on R2R in single-run setting. *pretraining-based methods. }
	\label{table:r2r_result}
	\vspace{-8pt}
\end{table*}

\subsubsection{Solutions for the Domain Gap}  
The domain gap brings critical challenges and researchers put forward methods to fill the gap between these two settings. 
Mobile robot navigation is considered a geometric problem, which requires the robot to sense the geometric shape of the environment in order to plan collision-free paths to reach the target. 
Obstacle avoidance is one of the most important challenges, and many methods~\cite{thorpe1988vision, muller2006off, francis2020long} have been put forward in previous work to achieve this. 
However, robot navigation in simulated tasks are regarded as a policy learning problem that learns a robust navigation policy from a starting position to the target in a complex environment with many possible routes. 
SLAM-based methods as in~\cite{gupta2017cognitive, zhang2017neural} contribute a lot to mapping and path planning, which is general for both simulated and real navigation. 
Deep learning shows its ability in processing images and learning policies for robotic control, which is widely applied in both settings. 
However, the usages of deep learning are different between simulated navigation and real navigation. 
In real-world navigation, the deep neural network is used to perceive RGB inputs~\cite{hadsell2009learning},  predict the future~\cite{finn2017deep} and learn the navigation policy~\cite{francis2020long}. 
However, due to the sampling inefficiency and the complex dynamic factors of the real-world environment, the policy is not robust enough. 
Some works~\cite{ha2018world, borenstein1990real} propose to model the environment and other works~\cite{francis2020long} adopt handcrafted rules to improve the robustness of the navigation policy. 
Data sampling is much more efficient in simulated environments. 
Most of the simulators render RGB and depth images in more than hundreds of frames per second (FPS), in which the fastest simulator, Habitat~\cite{savva2019habitat}, achieves 100,000 FPS. 
Fast data sampling enables learning with large batch size. 
Many works prove that a large training batch size leads to robustness in representations~\cite{malone2012dd, wani2020multion}. 
In spite of the rendered RGB and depth images, some simulated environments are able to provide semantic segmentation masks~\cite{eqa_matterport, xia2018gibson, savva2019habitat}. 
A more accurate simulator with few noise to facilitate training. 
With richer, noise-free data, researchers can apply a deeper neural network on navigation agents without worrying about overfitting. 
For example, Transformer~\cite{vaswani2017attention} is widely applied in navigation works in simulated environments due to its capability of feature representation while it is easily overfitting if it is trained on  noisy data. 

\begin{figure}[t]
	\centering
	\includegraphics[width=0.98\linewidth]{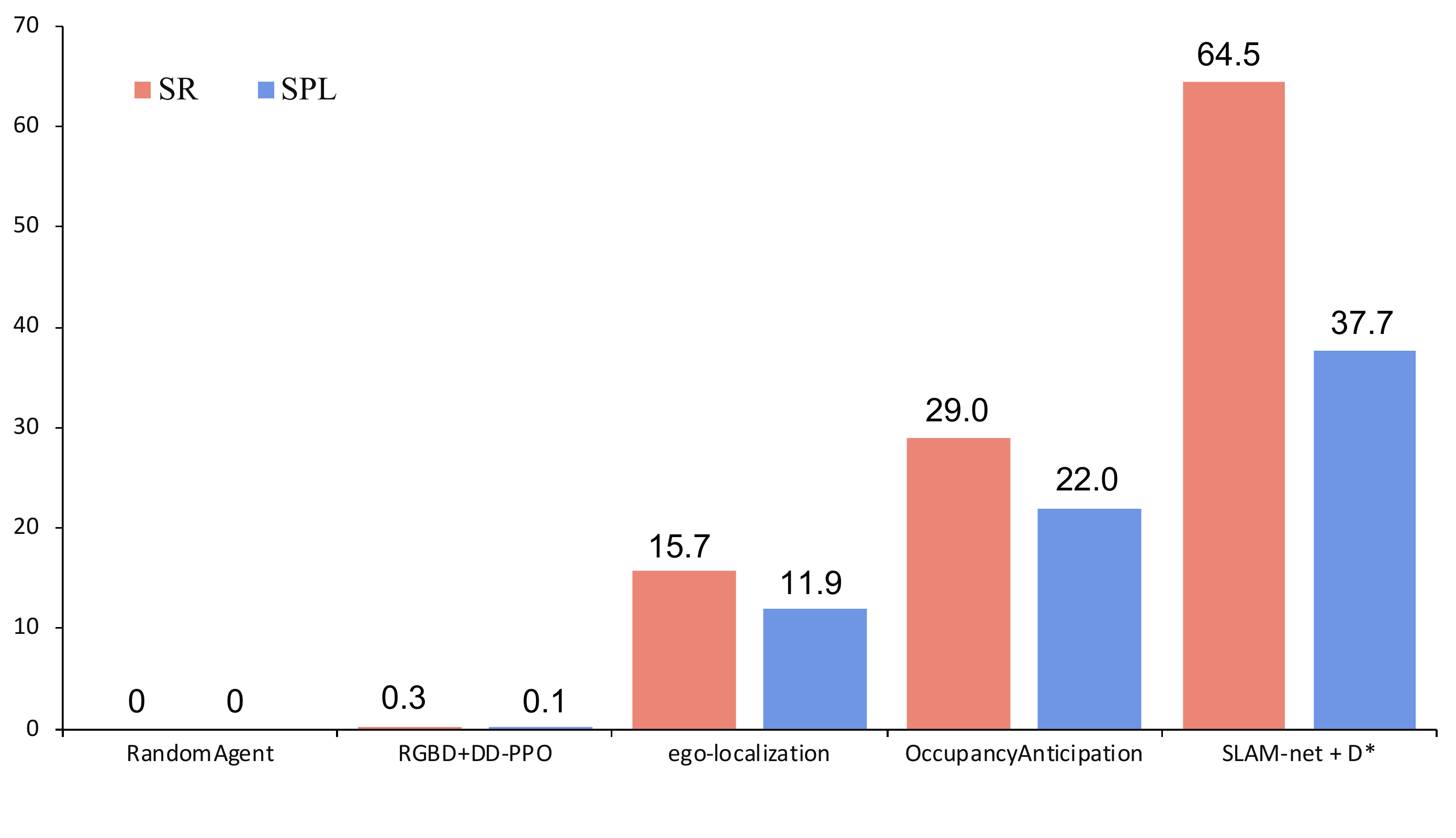}
	\vspace{-6pt}
	\caption{
	The performances of methods on the Habitat \emph{PointGoal} Challenge, including DD-PPO~\cite{wijmans2020dd}, ego-localization~\cite{datta2020integrating}, Occupancy Anticipation~\cite{ramakrishnan2020occupancy} and SLAM-net. }
	\label{fig:habitat_p}
	\vspace{-10pt}
\end{figure}

\subsubsection{Learning Efficiency} 
Many researchers focus on learning efficiency since data sampling in the real world is slow and expensive. 
Lobos-Tsunekawa \emph{et al.}~\cite{lobos-tsunekawa2018visual} propose a map-less visual navigation method for biped humanoid robots.
In this method, DDPG algorithm~\cite{lillicrap2016continuous} is used to extract information from color images, so as to derive motion commands. 
This method runs 20 ms on a physical robot, allowing its use in real-time applications. 
Bruce \emph{et al.}~\cite{bruce2017one} present a method for learning to navigate to a fixed goal on a mobile robot. 
By using an interactive replay of
a single traversal of the environment and stochastic environmental augmentation, Bruce \emph{et al.} demonstrates zero-shot transfer under real-world environmental variations without fine-tuning. 
To further improve the sampling efficiency, 
Pfeiffer~\emph{et al.}~\cite{pfeiffer2018reinforced} leverage prior
expert demonstrations for pre-training so that the training cost could be largely reduced in the fine-tuning process.

\subsection{Navigation Transferring}
Transfer learning is attracting rising attention in embodied navigation. 
The researchers are motivated from two aspects: 
1) learn a navigation agent that is able to perform accurate and efficient navigation in diverse domains and tasks; 
2) deploy an agent trained in a simulated environment in a real-world navigation robot. 

It is challenging to train a model to learn skills for navigating in different domains. 
Moreover, due to the large domain gap between simulated environments and the real-world environment, a well-performed navigation policy trained on a simulated environment cannot be easily transferred to the real-world environment. 
A lot of navigation tasks have been proposed to investigate different capabilities for navigation in diverse scenarios. 

In this section, we discuss the transfer learning in navigation from two different levels: 
1) task-level transferring; 
2) environment-level transferring, including sim-to-real transferring. The task-level transferring requires the agent to learn a policy that adapts to different input modalities or targets; the environment-level transferring requires the model to be invariant to different dynamics and transition functions. 

DisCoRL\cite{traor2019discorl} introduce a policy distillation method~\cite{rusu2016policy} to transfer a 2D navigation policy. 
In addition to the navigation policy, the vision and language embedding layer could also be transferred ~\cite{huang2019transferable}. 
Motivated by the success of meta-learning~\cite{finn2017model}, Dimension-variable skill transfer (DVST)~\cite{zhang2020map} obtains a meta-agent with deep reinforcement learning and then transfers the meta-skill to a robot with a different dimensional configuration using a method named dimension-variable skill transfer. 
Similarly, Li \emph{et al.}~\cite{li2020unsupervised} propose an unsupervised reinforcement learning method to learn transferable meta-skills. 
Zhu \emph{et al.}~\cite{zhu2020babywalk} decompose long navigation instructions into shorter ones, and thus enables the model to be easily transferred to navigation tasks with longer trajectories. 
Chaplot \emph{et al.}~\cite{chaplot2020embodied} propose a multi-task model that jointly learns multi-modal tasks, and transfers vision-language knowledge across the tasks. 
The model adopts a Dual-Attention unit to disentangle the vision knowledge and language knowledge and align them with each other. 

\begin{figure*}[t]
	\centering
	\includegraphics[width=0.99\linewidth]{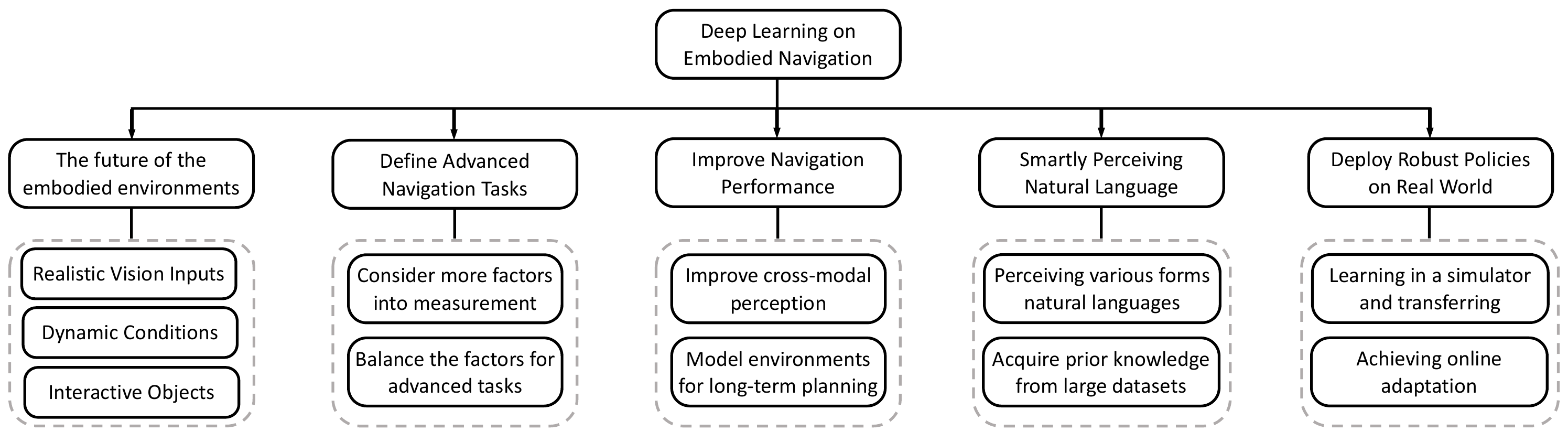}
	\caption{
	A summary of future directions for building an advanced robotic for real-world navigation. }
	\label{fig:future_directions}
	\vspace{-8pt}
\end{figure*}

Wang \emph{et al.}~\cite{wang2020environment} propose to learn environment-agnostic representations for the navigation policy enables the model to perform on both Vision-Language Navigation (VLN) and Navigation from Dialog History (NDH) tasks. 
Yan \emph{et al.}~\cite{yan2020multimodal} propose MVV-IN, a method that acquires transferable meta-skills with multi-modal inputs to cope with new tasks. 
Liu \emph{et al.}~\cite{liu2019lifelong} investigate on how to make robots fuse and transfer their experience so that they can effectively use prior knowledge and quickly adapt to new environments. 
Gordon \emph{et al.}~\cite{gordon2019splitnet} propose to decouple the visual perception and policy to facilitates transfer to new environments and tasks. 

Sim-real transferring have been well studied in the field of robotic control~\cite{tobin2017domain, rusu2017sim}. 
Sadeghi \emph{et al.}~\cite{sadeghi2017cad2rl} firstly propose a learning-based method, which trains a navigation agent entirely in a simulator and then transfers it into real-world environments without finetuning on any real images. 
Consequently, Yuan \emph{et al.}~\cite{yuan2019end} adopt a sim-real transfer strategy for learning navigation controllers using an end-to-end policy that maps raw pixels as visual input to control actions without any form of engineered feature extraction. 
Tai \emph{et al.}~\cite{tai2017virtual} train a robot in simulation with Asynchronous DDPG~\cite{lillicrap2016continuous} algorithm and directly deployed the
learned controller to a real robot for navigation transferring. 
Rusu \emph{et al.}~\cite{rusu2017sim} introduce a progressive network to transfer the learned policies from simulation to the real world.  
Similarly, adversarial feature adaptation methods~\cite{tzeng2017adversarial} is also applicable in sim-to-real policy transferring~\cite{zhu2019sim}. 
Sim-to-real transfer for deep reinforcement learning policies can be applied to complex navigation tasks~\cite{zhao2020sim}, including six-legged robots~\cite{qin2019sim}, robots for soccer competitions~\cite{bassani2020learning}, etc. 

\subsection{Summary}

In this section, 
We firstly compare the difference between simulated environments and real-world environments. 
Then, we reason about the domain gaps that cause the domain gaps. 
Finally, we introduce some transfer learning works in the navigation to give a promising direction to solve this problem.

\section{Future Directions}
\label{sec:future_direction}
Although extensive works have addressed the navigation problem from diverse aspects, current research progress is still far from real artificial intelligence. 
Also, current work cannot build a robust robot for real-world navigation. 
We summarize the challenges in solving embodied AI into these aspects: 
1) the functions and the performance are limited by the embodied environment;
2) the navigation problem is not well defined;
3) the performances of embodied AI agents in complex environments are still poor;
4) perceiving natural language is difficult to learn; 
5) hard to deploy a trained navigation policy to the real-world application. 
%
%

\noindent\textbf{Future Embodied Environments.} 
The advanced functions in the environment help the navigation model to obtain high-level abilities. 
For instance, compared to the early embodied environments, the large scene in the Matterport3D~\cite{chang2017matterport3d} firstly requires the navigation model to explore and memorize the complex room structure. The vision-language navigation benchmark~\cite{anderson2018vision} enables the agents to perceive natural language. 
The interactive embodied environment like AI2-THOR~\cite{kolve2017ai2} and iGibson~\cite{xia2020interactive} enable the agent to perform interactive actions. The agent learned in an interactive environment is able to move an object, put an object and open a door. 

An environment with more functions is the basis of learning a smart agent. 
An agent must be able to handle a dynamic environment when the objects in the rooms with evolving conditions. 
In stead of navigating within the navigable areas like~\cite{anderson2018vision, savva2019habitat}, we expect an agent to find possible roads within a room that has many obstacles. 
In addition, we need an interactive agent, which can pick up and put down objects, move chairs, and interact with human beings.
Other modes such as walking, running, and climbing also need to be considered if we want to build a robust navigator within a complex indoor environment. 

\noindent\textbf{Define Advanced Navigation Tasks. } 
Even though many embodied navigation tasks and navigation metrics have been proposed, what is a good navigation policy remains unclear.
This problem has two folds: 1) what factors have to be considered; 2) how to balance these factors. 
As we analysed in Sec.~\ref{sec:evaluation_metrics}, the accuracy and efficiency are two main factors to evaluate the performance of navigation. 
However, the importance of accuracy and efficiency are different among metrics. Optimal navigation policy varies according to different metrics.
In the interactive navigation task proposed by~\cite{shen2020igibson}, the performance of the agent is evaluated by a path efficiency score and a effort efficiency score. The interactive navigation task varies the score weights in evaluation to test if an agent performs well in different settings. However, it is still unclear how the weight of the factors affects the test results and what setting it is in real-world application. 

Moreover, in the questioning-answering settings like Help Anna~\cite{nguyen2019help}, or RMM~\cite{roman2020rmm}, the frequency of questioning or requesting from the agent are take into consideration. 
Current methods regard that the performance is lower if the agent ask for more information from human. 
Balancing the `cost' of asking questions and the `cost' of navigation is still a challenging problem. 
We hope that the community could publish more works discussing on how to evaluate the advanced navigation behavior or comparing the differences of navigation policies between an agent and a human. 

\noindent\textbf{Improve Navigation Performance.} 
Current navigation agents still perform poorly even in easy navigation tasks such as \emph{PointGoal} task and \emph{ObjectGoal} task, as shown in Fig.~\ref{fig:habitat_p} and Fig.~\ref{fig:habitat_o}. 
The state-of-the art model of the \emph{PointGoal} task performs 64.5\% on success rate (SR) and 37.7\% on SPL, while on the \emph{ObjectGoal} task, the best model performs 21.08\% on success rate (SR) and 8.38\% on SPL. 
As shown in Fig.~\ref{table:r2r_result}, the current state-of-the-art model~\cite{hong2021vln} in the vision-language navigation task performs 63\% in SR and 57\% in SPL while the human performance is 86\% in SR and 76\% in SPL. 
The navigation performances of current models are still far from human performance. 
Besides, existing models usually perform poorly on the challenging task of vision-and-language navigation. 
As shown in Tab.~\ref{table:r2r_result}, there is about 19\% performance gap between the state-of-the-art model and the human baseline. 
In the interactive dialog tasks~\cite{nguyen2019help, roman2020rmm, zhuyi:iccv2021}, the natural language used by agents to interact with oracle has many errors and is not fluent.
The baselines in the recently proposed tasks~\cite{shen2020igibson} can hardly complete the task. 
Moreover, the navigation robot in the real world cannot perform as well as in the simulated environment. 

Several promising directions, which are motivated by referring recent works, could tackle these problems. 
Transformer~\cite{vaswani2017attention} shows its capability in feature extraction and cross-modal fusion. 
Some works~\cite{hao2020towards, hong2021vln} build navigation models based on Transformer and achieve great success in vision-language navigation task, which reveals that Transformer structure is beneficial for cross-modal navigation policy. 
Chaplot \emph{et al.}~\cite{chaplot2020neural} build a neural SLAM module into a navigation model and train the model via hierarchical reinforcement learning. This model is able to learn the structure of the room and perform a robust low-level navigation policy in an environment with continuous state space. We suggest that model-based methods~\cite{ye2021auxiliary, zhu2020vision} and hierarchical reinforcement learning~\cite{jeni2007hierarchical} are the key to build a robust navigation model. 

\noindent\textbf{Smartly Perceiving Natural Language.}
Natural language is a complex modality for a robot to understand due to its diversity and complexity. 
At this moment, however, teach a navigation robot to learn to understand language requires a large amount of natural language annotations and each of them describes the semantics of a trajectory, a scene or a kind of behavior. The language annotations can be a word, a sentence, a question-answer pair or a dialogue, which are pretty expensive and labor-intensive.  

Even if we have sufficient language annotations for training a navigation robot, it is still challenging for the robot to correctly understand language instructions. 
For example, because there are many natural language variants for describing the same trajectory or scene, supervising an agent with trajectory-instruction pairs may led to severe overfitting. 
In addition, the skill of perceiving natural language needs prior knowledge. 
For example, ``find the forth chair in the living room'' requires the agent be able to count and ``navigate to bathroom safely'' requires the agent to turn smoothly and do not touch any objects. 
Some works~\cite{hao2020towards, hong2021vln} adopt pre-training methods to obtain a better language understanding skill with prior knowledge. 
Based on the success of these works, we believe that learning from other large-scale language datasets~\cite{yang2019xlnet, lee2019biobert, devlin2018bert} and transfer the prior knowledge might be a promising direction in solving the challenges in understanding natural language instructions. 


\noindent\textbf{Deploy Robust Policies on Real World.} 
Even though we have obtained a robust navigation policy in a simulated environment, how to deploy this policy to real-world still remains challenging. As demonstrated in Sec.~\ref{sec:sim-real}, three major differences cause the large sim-real domain gap: 1) observation; 2) action space; 3) environmental dynamics. 
Large sim-real domain gap hinders the direct deployment of the learned navigation policy to the real world. 
There are two directions in tackling this problem. 
One way is building a realistic simulator, including a realistic visual image rendering mechanism, advanced physical sensors, obstacle objects, dynamic simulation, simulation of robot components like wheels and gears, etc. 
However, such a realistic visual simulator is computation costly. 
Another way of solving sim-real deployment problem is achieving online adaptation by transfer learning or meta-reinforcement learning~\cite{levine2014learning, finn2017model}. 
These methods enable an agent to change its policy to adapt the environment. 
This method not only has high computational efficiency, but also has stronger adaptability when accidents happen. 



\section{Conclusion}
This paper presented a comprehensive survey on the embodied navigation scenario by summarizing hundreds of works. 
We thoroughly investigate the environments, tasks, and metrics to introduce the problem that the researchers are trying to solve. 
And we introduce hundreds of methods that solve these tasks in the embodied environments and compare their differences. 
Then we introduce the methods in the real-world environment and demonstrate how the large domain gap led to the drop in navigation performance. At last, we analyze the current problems that exist in the embodied navigation and give out four future directions to improve our community. 


\bibliographystyle{IEEEtran}
\bibliography{IEEEabrv, egbib.bib}


%






\vspace{-40pt}

\begin{IEEEbiography}[{\includegraphics[width=1in,height=1.2in,clip,keepaspectratio]{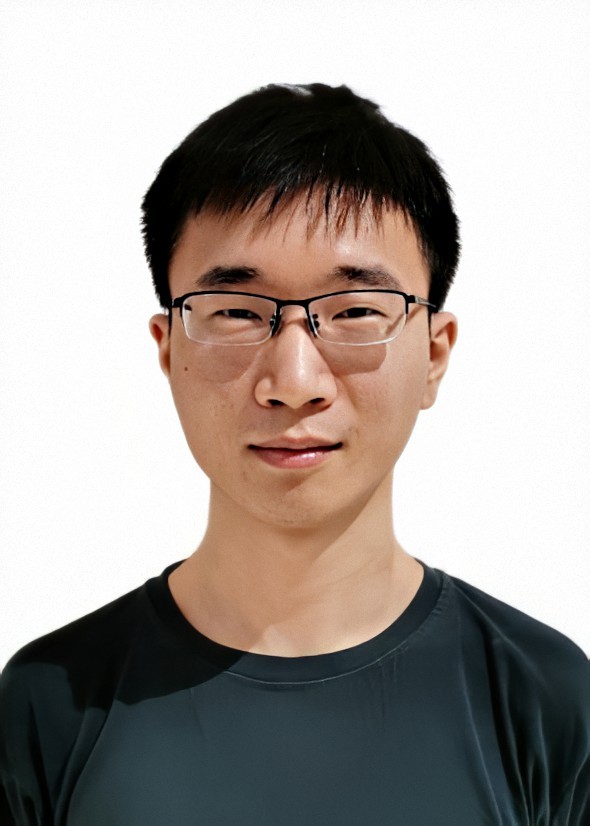}}]{Fengda Zhu}
received the bachelor’s degree in School of Software Engineering from Beihang University, Beijing, China, in 2017. He is currently pursuing the Ph.D. degree with the Faculty of Information Technology, Monash University under the supervision of Prof. Xiaojun Chang. His research interests include machine learning, deep learning and reinforcement learning.
\end{IEEEbiography}

\vspace{-50pt}

\begin{IEEEbiography}[{\includegraphics[width=1in,height=1.2in,clip,keepaspectratio]{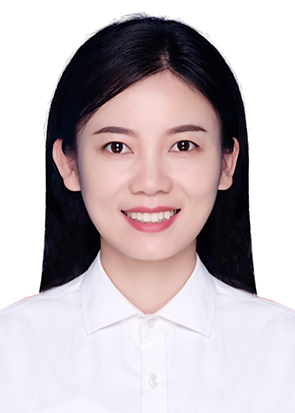}}]{Yi Zhu}
received the B.S. degree in software engineering from Sun Yat-sen University, Guangzhou,
China, in 2013. Since 2015,
she has been a Ph.D student in computer science with the School of Electronic, Electrical, and Communication Engineering, University of Chinese Academy of Sciences,
Beijing, China. Her current research interests include
object recognition, scene understanding, weakly supervised learning and visual reasoning.
\end{IEEEbiography}

\vspace{-50pt}

\begin{IEEEbiography}[{\includegraphics[width=1in,height=1.2in,clip,keepaspectratio]{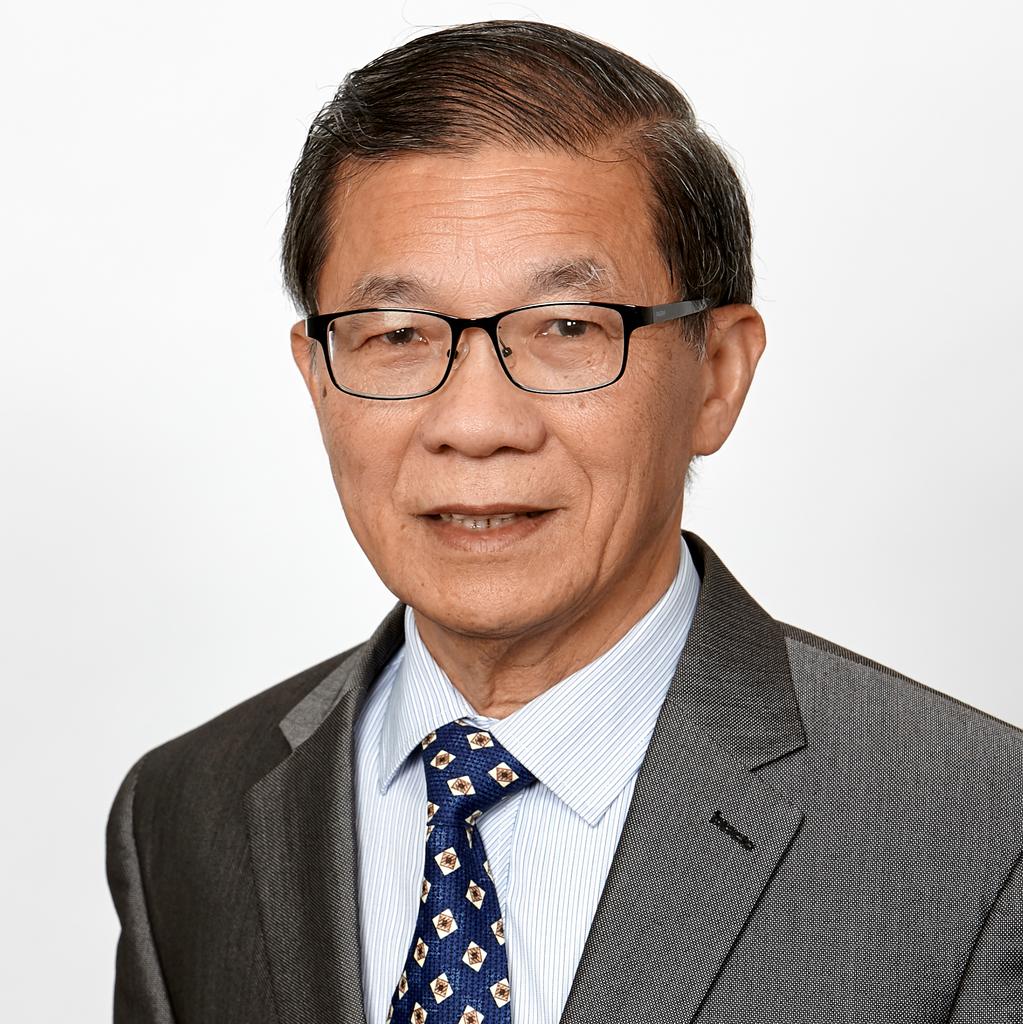}}]{Vincent Lee} is currently an Associate Professor at Machine learning and Deep Learning Discipline of the Department of Data Science and artificial Intelligence, Faculty of IT, Monash University, Australia. He is a senior member of IEEE USA. He received Australia Federal Government scholarship to pursue PhD from 1988 through to 1991 at The University of New Castle, NSW, in Australia. In 1973 to 1974, he was awarded a joint research scholarship by Ministry of Defence (Singapore) and Ministry of Defence (UK) for postgraduate study at Royal Air Force College, UK in aircraft electrical and instrument systems. He was a visiting academic to Tsinghua University in Beijing at the School of Economics and Management from Nov 2006 to March 2007; and was also a Visiting Professor to Information Communication Institute of Singapore from July 1994 through June 1995. 
\end{IEEEbiography}

\vspace{-40pt}

\begin{IEEEbiography}[{\includegraphics[width=1in,height=1.2in,clip,keepaspectratio]{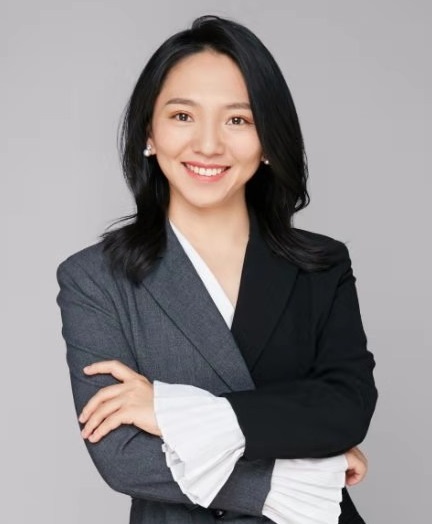}}]{Xiaodan Liang} is currently an Associate Professor at Sun Yat-sen University. She was a postdoc researcher in the machine learning department at Carnegie Mellon University, working with Prof. Eric Xing, from 2016 to 2018. She received her PhD degree from Sun Yat-sen University in 2016, advised by Liang Lin. She has published several cutting-edge projects on human-related analysis, including human parsing, pedestrian detection and instance segmentation, 2D/3D human pose estimation and activity recognition.
\end{IEEEbiography}

\vspace{-40pt}

\begin{IEEEbiography}[{\includegraphics[width=1in,height=1.2in,clip,keepaspectratio]{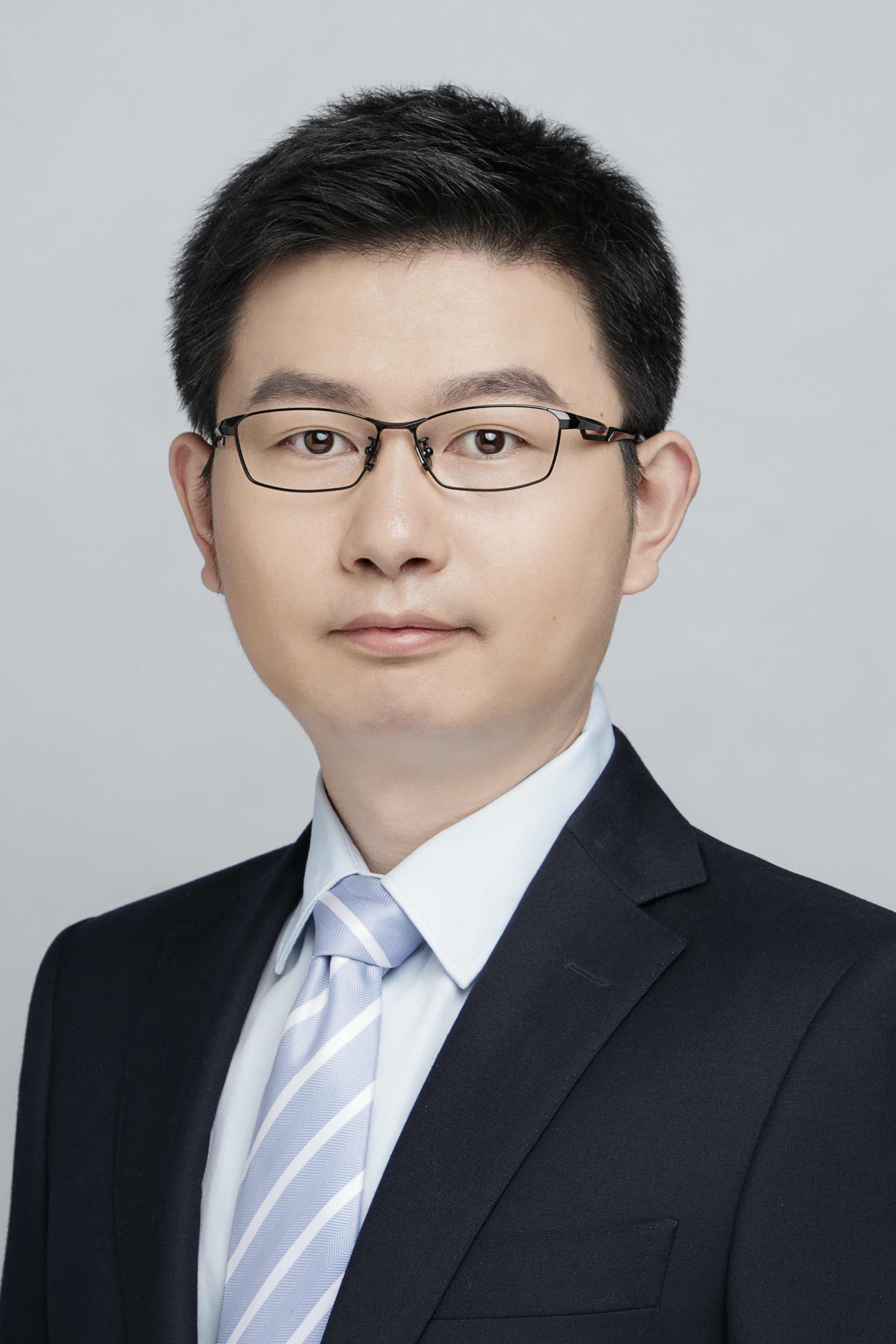}}]{Xiaojun Chang}
is currently an Associate Professor with School of Computing Technologies, RMIT University, Australia. Before joining RMIT, he was a Senior Lecturer with the Faculty of Information Technology, Monash University Clayton Campus, Clayton, VIC, Australia. He is also with the Monash University Centre for Data Science. He was a Post-Doctoral Research Associate with the School of Computer Science, Carnegie Mellon University, Pittsburgh, PA, USA, working with Prof. A. Hauptmann. He has spent most of the time working on exploring multiple signals (visual, acoustic, and textual) for automatic content analysis in unconstrained or surveillance videos.

Dr. Chang is an ARC Discovery Early Career Researcher Award (DECRA) Fellow from 2019 to 2021. He has achieved top performance in various international competitions, such as TRECVID MED, TRECVID SIN, and TRECVID AVS.
\end{IEEEbiography}

\end{document}